\documentclass[10pt]{IEEEtran}
\IEEEoverridecommandlockouts
\def\BibTeX{{\rm B\kern-.05em{\sc i\kern-.025em b}\kern-.08em
    T\kern-.1667em\lower.7ex\hbox{E}\kern-.125emX}}

\title{\sysdgg/: An End-to-End Document Structure Generator\\
\thanks{CZ and the DS3Lab gratefully acknowledge the support from the Swiss State Secretariat for Education, Research and Innovation (SERI) under contract number MB22.00036 (for European Research Council (ERC) Starting Grant TRIDENT 101042665), the Swiss National Science Foundation (Project Number 200021, 184628, and 197485), Innosuisse/SNF BRIDGE Discovery (Project Number 40B2-0 187132), European Union Horizon 2020 Research and Innovation Programme (DAPHNE, 957407), Botnar Research Centre for Child Health, Swiss Data Science Center, Alibaba, Cisco, eBay, Google Focused Research Awards, Kuaishou Inc., Oracle Labs, Zurich Insurance, and the Department of Computer Science at ETH Zurich.}
}

\author{
\IEEEauthorblockN{Johannes Rausch\IEEEauthorrefmark{2}, Gentiana Rashiti\IEEEauthorrefmark{2}, Maxim Gusev\IEEEauthorrefmark{2}, Ce Zhang\IEEEauthorrefmark{2}, Stefan Feuerriegel\IEEEauthorrefmark{3}}
\\\IEEEauthorblockA{\IEEEauthorrefmark{2}Department of Computer Science, ETH Zurich
}%
\IEEEauthorblockA{\IEEEauthorrefmark{3}Munich Center for Machine Learning, LMU Munich
\\johannes.rausch@inf.ethz.ch, rashitig@student.ethz.ch, gusevm@student.ethz.ch, ce.zhang@inf.ethz.ch, feuerriegel@lmu.de}
}

\usepackage{hyperref}
\usepackage{textcomp}
\usepackage{booktabs}

\usepackage{listings}
\usepackage{todonotes}
\usepackage{algorithm}

\usepackage{multirow}
\usepackage{makecell}

\usepackage[export]{adjustbox}

\usepackage{xspace}

\newcommand{\naive}{}
\def\naive/{na\"{\i}ve}

\newcommand{\sysdgg}{}
\newcommand{\syslong}{}
\newcommand{\datasetAD}{}
\newcommand{\datasetADtarget}{}
\newcommand{\datasetADweak}{}
\newcommand{\datasetEP}{}
\def\datasetAD/{arXivdocs}
\def\datasetADtarget/{arXivdocs-target}
\def\datasetADweak/{arXivdocs-weak}
\def\datasetEP/{\mbox{E-Periodica}}
\def\sysdgg/{\textsf{DSG}}
\def\syslong/{Document Structure Generator}

\usepackage{amsmath,amssymb,amsfonts}
\usepackage{cleveref}
\usepackage{siunitx}
\usepackage{subcaption}

\usepackage{multicol}
\usepackage{paralist}
\usepackage{graphicx}

\usepackage{placeins}

\usepackage{multibib}

\usepackage{fancyvrb,xcolor}
\definecolor{codegreen}{rgb}{0,0.6,0}
\definecolor{codegray}{rgb}{0.5,0.5,0.5}
\definecolor{codepurple}{rgb}{0.58,0,0.82}
\definecolor{backcolour}{rgb}{0.95,0.95,0.92}

\usepackage{pifont}%
\definecolor{ForestGreen}{RGB}{34,139,34}
\definecolor{BrickRed}{RGB}{170,74,68}

\newcommand{\cmark}{\textcolor{ForestGreen}{\ding{51}}}%
\newcommand{\xmark}{\textcolor{BrickRed}{\ding{55}}}%

\lstdefinestyle{pythonstyle}{
    backgroundcolor=\color{backcolour},   
    commentstyle=\color{codegreen},
    keywordstyle=\color{magenta},
    stringstyle=\color{codepurple},
    basicstyle=\ttfamily\footnotesize,
    breakatwhitespace=false,         
    breaklines=true,                 
    captionpos=b,                    
    keepspaces=true,                 
    numbers=none,                    
    showspaces=false,                
    showstringspaces=false,
    showtabs=false,                  
    tabsize=2
}

\begin{document}

\maketitle

\begin{abstract}
Information in industry, research, and the public sector is widely stored as rendered documents (e.g., PDF files, scans). Hence, to enable downstream tasks, systems are needed that map rendered documents onto a structured hierarchical format. However, existing systems for this task are limited by heuristics and are \emph{not} end-to-end trainable. In this work, we introduce the \emph{\syslong/} (\sysdgg/), a novel system for document parsing that is fully end-to-end trainable. \sysdgg/ combines a deep neural network for parsing (i)~entities in documents (e.g., figures, text blocks, headers, etc.) and (ii)~relations that capture the sequence and nested structure between entities. Unlike existing systems that rely on heuristics, our \sysdgg/ is trained end-to-end, making it effective and flexible for real-world applications. We further contribute a new, large-scale dataset called \datasetEP/ comprising real-world magazines with complex document structures for evaluation. Our results demonstrate that our \sysdgg/ outperforms commercial OCR tools and, on top of that, achieves state-of-the-art performance. To the best of our knowledge, our \sysdgg/ system is the first end-to-end trainable system for hierarchical document parsing.
\end{abstract}


\vspace{-0.35cm}
\section{Introduction}
\vspace{-0.15cm}

Large amounts of information are generated daily in industry, research, and the public sector. Yet, such data is typically stored as document renderings (e.g., PDF files, scans) and not as structured hierarchical formats \cite{johnsonPdfStatisticsUniverse2018}. This is a crucial hurdle for practice. On the one hand, structured formats as opposed to document renderings are needed for efficient storage in databases, as the latter requires standardized formats \cite{johnsonNativeXMLDatabase2003,cliftonDesignDocumentDatabase2000}. On the other hand, document renderings cannot be processed in downstream tasks since downstream tasks commonly require documents that are in a parsable format. Examples are query and retrieval \cite{cheQueryOptimizationXML2006,cafarellaWebTablesExploringPower2008,wilkinsonEffectiveRetrievalStructured1994,liIndexingQueryingXML2001,manabeExtractingLogicalHierarchical2015} and knowledge base construction \cite{wuFonduerKnowledgeBase2018}. To this end, there is a direct need in practice \cite{binmakhashenDocumentLayoutAnalysis2019} for systems that map document renderings onto a structured hierarchical format.

\begin{figure}[tbh!]
    \centering
    \includegraphics[width=\linewidth]{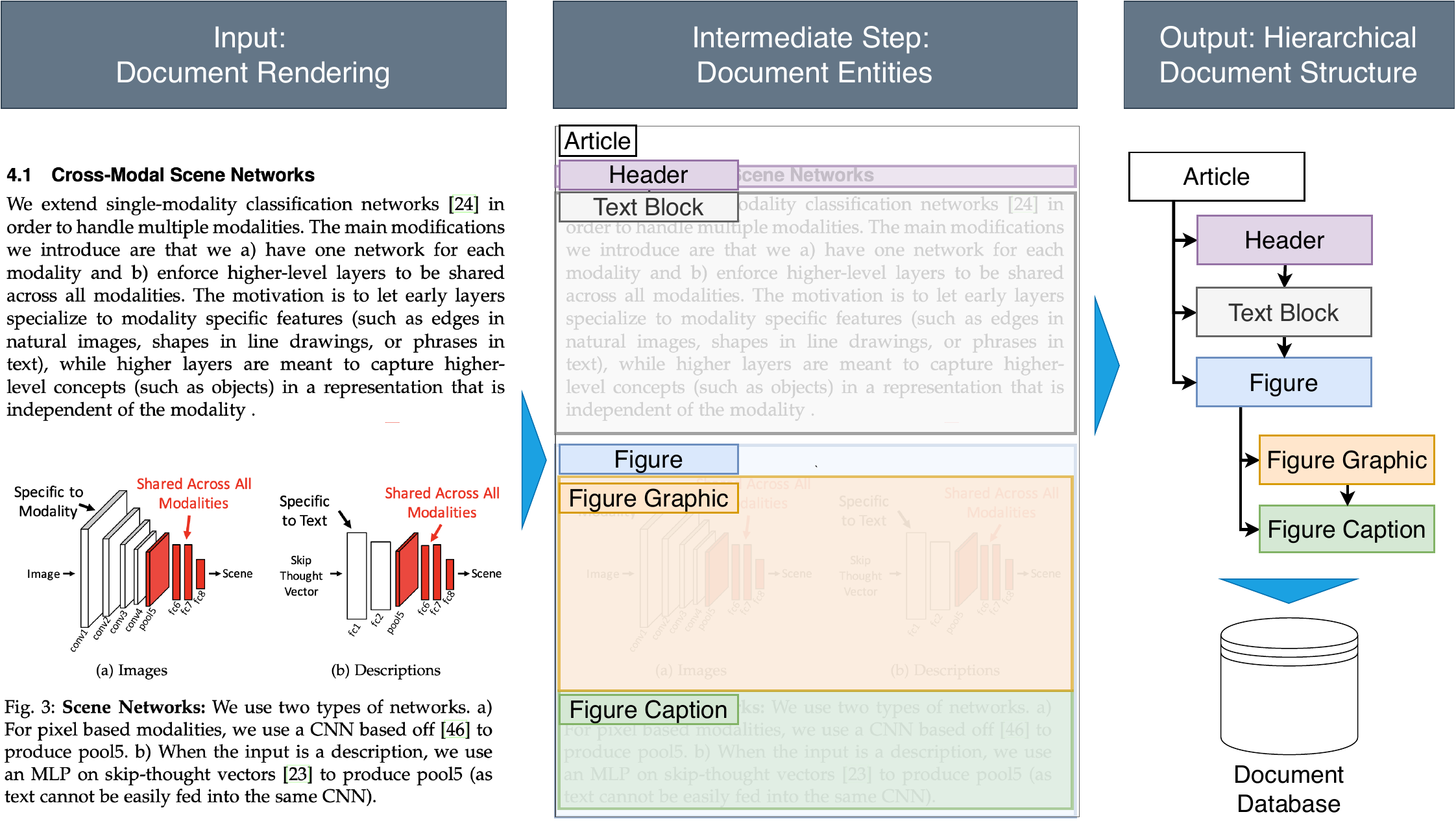}
\vspace{-0.25cm}
    \caption{The task of generating document structures by identifying (i)~entities within documents and (ii)~relations describing the hierarchical structure.}
    \label{fig:hierarchical_parsing_workflow}
\end{figure}

The task of document structure parsing refers to the generation of a hierarchical document structure, given a document rendering as input (see Fig.~\ref{fig:hierarchical_parsing_workflow}). For this, the textual contents must be extracted from the document rendering while preserving the semantic and hierarchical structure of the source files. To achieve this, state-of-the-art systems \cite{rauschDocParserHierarchicalDocument2021} typically first detect all document entities (e.g., figures, text blocks, headers) and subsequently infer the hierarchical relationships between entities (e.g., their sequence and nested structure) to form hierarchical document structures. Yet, the task is challenging due to the complex, nested structures in real-world documents.

A standard solution to map document renderings onto parsable documents are optical character recognition (OCR) systems. Current OCR systems are highly effective in retrieving word-level textual contents from rendered documents \cite{breuelHighPerformanceText2017a}. However, OCR systems generally focus only on the textual contents but struggle with inferring the hierarchical structure. OCR systems generally rely upon a prior step for parsing the document structure, yet which is still very challenging and error-prone \cite{binmakhashenDocumentLayoutAnalysis2019}. As a consequence, OCR systems suffer from large errors, especially if errors in the step for parsing the structure parsing occur \cite{zhuDocBedMultiStageOCR2022b}. To address this, earlier research focused on custom systems for parsing specific entities in documents such as table structures \cite{liTableBankBenchmarkDataset2020a,smockPubTables1MComprehensiveTable2022} but without parsing the complete hierarchical structure in documents. Even other research aimed at identifying document entities \cite{antonacopoulosRealisticDatasetPerformance2009a,zhongPubLayNetLargestDataset2019a,appalarajuDocFormerEndtoEndTransformer2021a}, but without actually generating hierarchical document structures. Only one work is tailored to generate hierarchical document structures from document renderings \cite{rauschDocParserHierarchicalDocument2021}. Yet, this work is based on heuristics and is thus \emph{not} end-to-end trainable, which is why its flexibility is limited. To the best of our knowledge, there is no system for parsing hierarchical document structures that is \emph{end-to-end} trainable.

\textbf{Our \sysdgg/ system:}\footnote{\label{opensource_link}Codes for our system, online supplements, and dataset are publicly available at \url{https://github.com/j-rausch/DSG}.} We develop the \emph{\syslong/} (\sysdgg/), a novel system for generating hierarchical document structures from document renderings where the system is fully \emph{end-to-end trainable}. Our \sysdgg/ builds upon a deep neural network for parsing (i)~entities in documents (e.g., figures, text blocks, headers, etc.) and (ii)~relations that capture the sequence and nested structure between entities. In contrast to existing systems for generating document structures, our \sysdgg/ uses a trainable component for classifying relations and thereby circumvents the use of heuristics. As a result, our \sysdgg/ predicts entire document structures and is thus fully end-to-end trainable.  This makes our system highly flexible for handling a variety of documents that can arise in practice. Finally, our \sysdgg/ has a custom conversion engine to generate structured document output files in hOCR markup language, which allows for seamless integration into existing document storage and processing workflows.

We further contribute a novel, large-scale dataset for generating and evaluating hierarchical document structures called \datasetEP/. \datasetEP/ is based on real-world magazines from different source languages (e.g., English, German, French, Italian). We manually annotated the hierarchical document structure for several hundred magazine pages. Overall, \datasetEP/ contains $542$ documents with more than $11,000$ annotated entities. Thereby, we extend over previous datasets that have been primarily limited to scientific articles \cite{rauschDocParserHierarchicalDocument2021}. However, a limitation of scientific articles is that they follow a fairly similar structure, while magazines are characterized by large heterogeneity in their presentation and thus complex document structures. Hence, \datasetEP/ provides a novel and challenging, real-world setting for evaluation.

Our \textbf{main contributions} are as follows:
\begin{enumerate}
\item We develop a novel system for generating hierarchical document structures from document renderings called \sysdgg/. To the best of our knowledge, our \sysdgg/ is the first system for this task that is end-to-end trainable.
\item We contribute a novel, large-scale dataset called \datasetEP/ with manual annotations for evaluation. 
\item We show that our \sysdgg/ system achieves state-of-the-art performance. We further demonstrate the effectiveness of end-to-end training.
\end{enumerate}

\vspace{-0.35cm}
\section{Related work}
\vspace{-0.15cm}
\label{sec:related_work}

\textbf{Document structure parsing:} Existing systems have two shortcomings in that they are either (i)~limited to entity recognition and thus do \emph{not} generate hierarchical structures, or (i)~based on heuristics and thus \emph{not} end-to-end trainable. We provide a detailed overview in the following.

\underline{OCR systems:} Extracting text from document images has been extensively studied as part of OCR systems \cite{schaferACLAnthologySearchbench2011,Schafer2012}. As such, the works extract textual content, but \emph{not} the hierarchical document structure, which is the objective of our research.

\underline{Entity-specific parsers:} Several systems focus on specific semantic entities, namely, table detection and table structure parsing. In table detection, the task is to predict the bounding boxes of tables within document renderings, rather than generating the actual table structures \cite{Yildiz2005,Wang2004,gilaniTableDetectionUsing2017,liTableBankBenchmarkDataset2020a,Siddiqui2018}. In table structure parsing, the aim is to recognize the structure (e.g., rows, cells) in tables \cite{smockPubTables1MComprehensiveTable2022, liTableBankBenchmarkDataset2020a}. Here, the input is provided either as text \cite{Kieninger1998,pivkTransformingArbitraryTables2007} or through document renderings \cite{schreiberDeepDeSRTDeepLearning2017a}. However, the works are limited to a single entity (tables). Hence, these works cannot identify other entities and thus \emph{not} the full document structure. Others works (e.g., \cite{zhongPubLayNetLargestDataset2019a,appalarajuDocFormerEndtoEndTransformer2021a}.) perform entity detection in documents by locating specific elements, but again \emph{without} extracting hierarchical structures. Even others focused only on the segmentation of individual lines \cite{josephReviewVariousLine2021}.

\underline{Hierarchical document parsers:} One work \cite{maHRDocDatasetBaseline2023} classifies lines and their immediate parent lines but using third-party OCR tools for that leads to error propagation throughout the system. More importantly, the work is limited to a few, high-level entities (e.g., table structures, lists, and sub-figures are missing) and most of the hierarchical structure is lost (e.g., orderings). Hence, this work fails to generate comprehensive document structures as required in our task. 

Closest to our work is a system called DocParser \cite{rauschDocParserHierarchicalDocument2021}, which is specifically designed to capture \emph{hierarchical} document structures. DocParser consists of five components (image conversion, entity detection, relation classification, structure-based refinement, and scalable weak supervision) in order to generate both entities and relations. However, in DocParser, relations are detected based on manual heuristics and are \emph{not} trainable. Hence, to the best of our knowledge, systems for hierarchical document parsing that are \emph{end-to-end} trainable are lacking.

\textbf{Scene graph generation:} Scene graph generation is a computer vision task that combines the entity detection vision task with an additional relation classification \cite{changComprehensiveSurveyScene2023}. Many recent methods for scene graph generation are based on two-stage training procedures where the detection components build upon Faster R-CNN \cite{zellersNeuralMotifsScene2018,khandelwalSegmentationgroundedSceneGraph2021,tangLearningComposeDynamic2019}. However, systems for scene graph generation are predominantly used to parse real-world images and, to the best of our knowledge, have not yet been adapted to document structure parsing. This is our contribution.

\textbf{Research gap:} Existing systems for generating hierarchical document structures are based on heuristics and are thus \emph{not} end-to-end-trainable, which limits their flexibility. As a remedy, we develop our \sysdgg/, the first system for hierarchical document parsing that is \emph{end-to-end} trainable.

\vspace{-0.35cm}
\section{Problem description}
\vspace{-0.15cm}

\textbf{Objective:} The objective of our system is to generate \emph{hierarchical} document structures from document renderings (e.g., PDF files, scanned images). Formally, the input is given by document renderings $D_1, \ldots, D_n$. The outputs are hierarchically-structured documents given by pairs $(H_1, T_1), \dots, (H_n, T_n)$, where $H_i$, $i = 1\ldots, n$, captures the hierarchical structure and $T_i$, $i = 1\ldots, n$, the texts. The hierarchical structure is defined by a set of (i)~entities in the documents (e.g., figures, text blocks, headers, etc.) and (ii)~relations that capture the sequence and nested structure between entities. Formally, entities are given by $E_j$, $j = 1, \ldots, m$, and relations by $R_j$, $j = 1, \ldots, k$. Both are defined below. The above task is thus analogous to earlier research on parsing hierarchical document structures (e.g., \cite{rauschDocParserHierarchicalDocument2021}).

\textbf{Entities:} Entities capture the different structural elements in documents, such as figures, tables, captions, text blocks, etc. Each entity $E_j$, $j = 1, \ldots, m$, is described by three attributes: (1)~a semantic category $c_j \in \mathcal{C} = \{C_1, \ldots, C_l\}$ (e.g., whether it is a figure, table, header, etc.); (2)~a rectangular bounding box $B_j$ in the document rendering, defined by the $x$- and $y$- coordinates of corner points of the bounding box; and (3)~a confidence score $P_j$ that accompanies the prediction of the semantic category $c_j$.

\textbf{Relations:} Relations capture the nested structure among the entities. Relations $R_j$, $j = 1, \ldots, k$are defined by triples $(E_{\text{subj}}, E_{\text{obj}}, \Psi)$ consisting of a subject $E_{\text{subj}}$, an object $E_{\text{obj}}$, and a relation type $\Psi \in \{ \mathit{parent\_of},  \mathit{followed\_by}, \text{null} \}$. Furthermore, the relations $R_j$ are associated with a confidence score $P_j^{\Psi}$ for the predicted relation type $\Psi$. 

\textbf{hOCR output:} As additional output, hierarchically-structured text documents given by pairs $(H_1, T_1), \dots, (H_n, T_n)$, where $T_i$, $i = 1\ldots, n$, captures the texts should further be provided in the standardized hOCR format \cite{breuelHOCRMicroformatOCR2007} to facilitate downstream processing tasks. hOCR is a markup language for representing and storing structured documents in a unified format \cite{breuelHOCRMicroformatOCR2007}. Hence, this should ensure that the output can be directly used by common tools for document processing and storage workflows that are widespread in practice. 

\vspace{-0.35cm}
\section{Our \sysdgg/ system}
\vspace{-0.15cm}

\begin{figure}[tbh!]
    \centering
    \includegraphics[width=\linewidth]{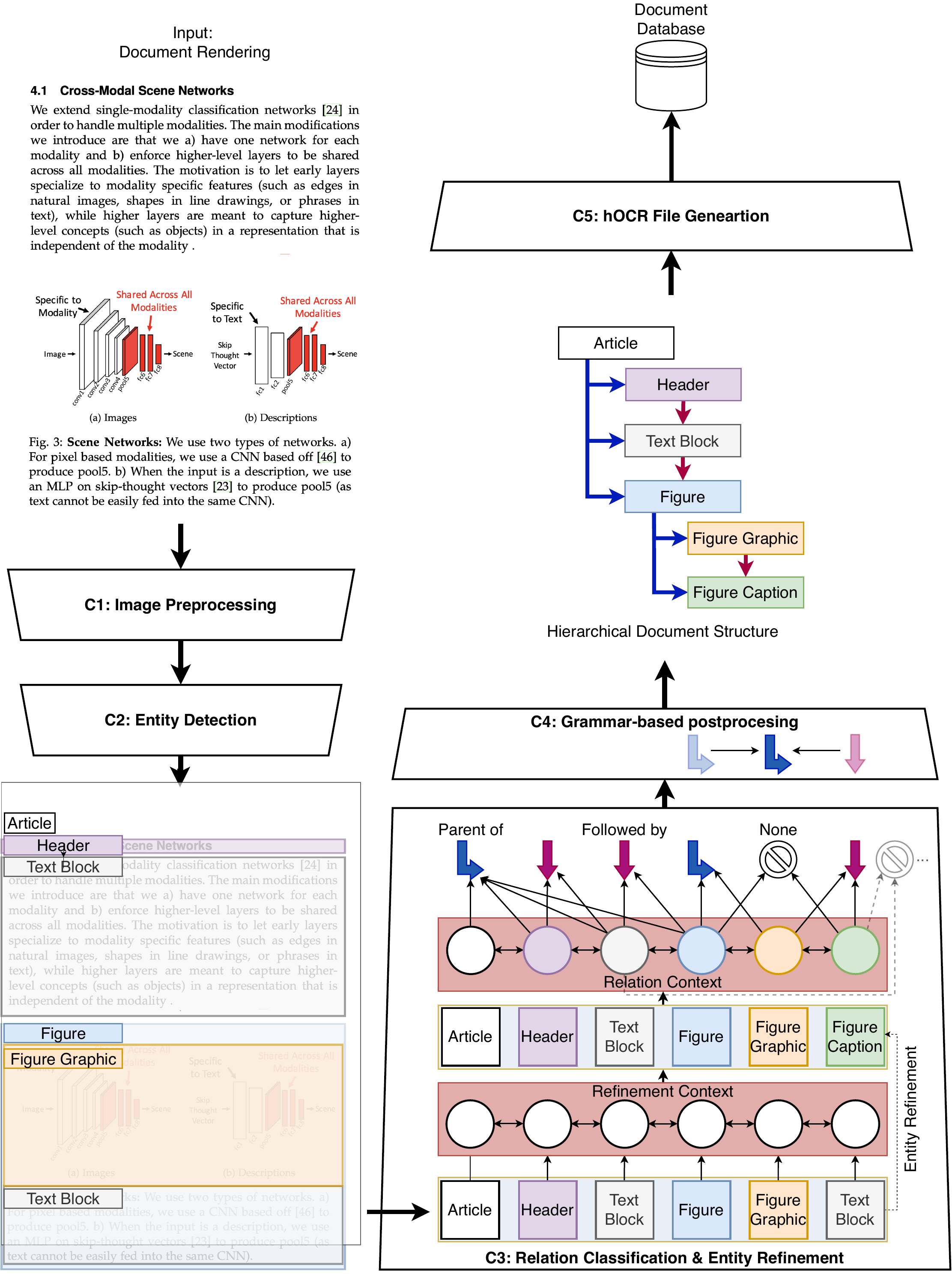}
    \caption{Overview of our \sysdgg/ system.}
\vspace{-0.25cm}
    \label{fig:dgg_system_overview}
\end{figure}

\textbf{Overview:} An overview of our \sysdgg/ system is shown in Fig.~\ref{fig:dgg_system_overview}. The objective of our \sysdgg/ is to generate hierarchical document structures from document renderings in an end-to-end trainable setup. For this, our system builds upon a deep neural network that consists of fully trainable components to parse both entities and subsequently the relations that represent the hierarchical structures. Our system processes documents along five components: (\textbf{C1})~image preprocessing, (\textbf{C2})~entity detection, (\textbf{C3})~relation classification and entity refinement, (\textbf{C4})~grammar-based postprocessing, and (\textbf{C5})~hOCR conversion engine. The components are described in the following.

\vspace{-0.35cm}
\subsection{Image preprocessing (C1)}
\vspace{-0.15cm}

Our system processes all input documents as rendered images. For source formats such as PDF, we first generate images for each document page, which are then used as input to our system.

Images are resized bilinearly so that their smallest side has a maximum size of $\phi_s^\mathrm{max}$. If the longest side exceeds a predefined maximum size of $\phi_l^{\mathrm{max}}$ after this step, the images are resized so that their longest side length is $\phi_l^{\mathrm{max}}$. During training, the image size can be varied for the purpose of data augmentation.\footnote{In computer vision, it is common to perform additional data augmentations through image mirroring or rotation operations, which are commonly applied to facilitate training and system performance. However, we avoid such data augmentations in our work because the hierarchical document structures are sensitive to the original document geometry (e.g. left-to-right reading orders).} For this, $\phi_s^\mathrm{max}$ and $\phi_l^\mathrm{max}$ are randomly chosen from a set of different sizes. The images are then normalized following the procedure in \cite{wu2019detectron2}. Specifically, we subtract the mean channel-wise pixel values of the underlying pre-training dataset \cite{dengImageNetLargescaleHierarchical2009} from the inputs.

\vspace{-0.35cm}
\subsection{Entity detection (C2)}
\vspace{-0.15cm}

The second component builds upon a Faster R-CNN architecture \cite{renFasterRCNNRealTime2015} for entity detection. Here, visual feature maps on different scales are extracted via a convolutional neural network \cite{heDeepResidualLearning2016,Lin2017}. The visual feature maps are then passed on to another network component, called region proposal network (RPN), which generates a set of rectangular candidate entity region proposals in the image. For each of the region proposals, a category prediction network is applied to predict the semantic category $c'_j$ of an entity $E_j$. If the confidence score $P'_j$ of a candidate region surpasses a predefined threshold, it is accepted as an entity (and discarded otherwise). Subsequently, an additional neural network is used to predict the size and position of the initial rectangular region $B_j$ (based on the rectangular candidate entity region proposals from the RPN). Afterward, the entities are passed on to component~C3, which is responsible for the relation classification.

\vspace{-0.35cm}
\subsection{Relation classification and entity refinement (C3)}
\vspace{-0.15cm}

The relation classification in \sysdgg/ builds upon the neural motifs architecture \cite{zellersNeuralMotifsScene2018}. This architecture extends the entity detection architecture with two additional neural network heads. Concretely, the detected entities are passed on to neural network heads for relation classification and entity refinement. In the following, we refer to these two heads as the relation head and the refinement head, respectively. 

Both the relation head and the refinement head build on bidirectional long short-term memory (LSTM) networks that take the entities from component C2 as input. Both proceed in slightly different ways. The relation head is fed with pairs of subject entity and object entity, $(E_{\text{subj}}, E_{\text{obj}})$, to classify if they form a relation triple $(E_{\text{subj}}, E_{\text{obj}}, \Psi)$ of type $\Psi \in \{ \mathit{parent\_of}, \mathit{followed\_by}, \text{null} \}$. The refinement head maps the categorical labels $c'_j$ and their confidence scores $P'_j$ from component C2 onto refined categories $c_j$ and confidence scores $P_j$ by taking into account the contextual information of all predicted entities. Formally, both relation head and refinement head are implemented as follows:
\begin{enumerate}[(i)]
\item \emph{Relation head:} The relation classification returns a confidence score $P_j^{\Psi}$ for all considered entity pairs $(E_{\text{subj}}, E_{\text{obj}})$. If the respective confidence score exceeds a predefined threshold $\tau$, a relation $R_j$ of type $\Psi \in \{ \mathit{parent\_of},  \mathit{followed\_by}, \text{null} \}$ is accepted. Here, the relation type $\Psi = \text{null}$ is used to indicate the absence of a hierarchical relation.
\item \emph{Refinement head:} The refinement head is fed with additional features $\rho^{\text{ref\_in}}_{i}$, $i \in \{ \text{vis}, \text{cat}, \text{pos} \}$, as follows.  First, $\rho^{\text{ref\_in}}_\text{vis}$ refers to the visual feature map that is extracted by the underlying Faster R-CNN architecture and corresponds to the image region of the entity in the rendered document. Second, $\rho^{\text{ref\_in}}_\text{cat}$ is a category embedding, which is based 
on a pre-trained word embedding dictionary and is selected according to the predicted semantic category of the entity \cite{penningtonGloveGlobalVectors2014}. Third, $\rho^{\text{ref\_in}}_\text{pos}$ is a positional embedding to represent the size and location of the entity bounding box. Specifically, the positional embedding incorporates the width, height, and location of the bound box $B_j$. We refer later refer to the features $\rho^{\text{ref\_in}}_{i}$, $i \in \{ \text{vis}, \text{cat}, \text{pos} \}$ as \emph{refinement context input features}.

The refinement context input features are passed to the LSTM from the refinement head and, subsequently, a fully connected layer to produce so-called \emph{refinement context output features} $\rho^{\text{ref\_out}}$. These features are then used to predict the refined entity categories $c_j$.

Subsequently, the relation head is fed with three \emph{relation context input features} $\rho^{\text{rel\_in}}_{i}$, $i \in \{ \text{vis}, \text{cat}, \text{ref} \}$ as follows. First, $\rho^{\text{rel\_in}}_\text{vis}$ is visual feature map,  identical to that in $\rho^{\text{ref\_in}}_\text{vis}$. Second, $\rho^{\text{rel\_in}}_\text{cat}$ is the category embedding, analogous to the category embedding $\rho^{\text{ref\_in}}_\text{cat}$ to represent the refined entity category. Third, $\rho^{\text{rel\_in}}_\text{ref}$ are the refinement context output features, i.e., $\rho^{\text{rel\_in}}_\text{ref} = \rho^{\text{ref\_out}}$. 

Next, pairs of two entities are processed by the relation head. Crucially, unlike existing systems such as DocParser, this step is fully trainable. For this, the relation head forms three so-called \emph{pair-wise features} $\rho^{\text{pair}}_1$, $\rho^{\text{pair}}_2$, and $\rho^{\text{pair}}_3$ as specified in the following. Later, the pair-wise features are used to predict confidence scores $P_j^{\Psi}$ for all considered entity pairs $(E_{\text{subj}}, E_{\text{obj}})$. Specifically, the pair-wise features are: First,$\rho^{\text{pair}}_1$ is the visual feature map that is extracted by the Faster R-CNN architecture. It corresponds to the image region of the entity pair $B_{\text{pair}}=\text{Union}(E_{\text{subj}}, E_{\text{obj}})$. Second, $\rho^{\text{pair}}_2$ is formed for entity-entity pairs $(E_{\text{subj}}, E_{\text{obj}})$ by concatenating the respective refinement context output features $(\rho^{\text{rel\_out}}_\text{subj}, \rho^{\text{rel\_out}}_\text{obj})$. Third, a frequency bias term $\rho^{\text{pair}}_3$ is calculated from the refined categories $c$ of the pair-wise considered entities. The frequency bias term is based on the empirical distribution over relations $(E_{\text{subj}}, E_{\text{obj}}, \Psi)$ in the training set. The frequency bias term thus reflects that, for certain pairings of entity categories $(c_\text{subj},c_\text{obj})$, relation types $\Psi \in \{ \mathit{parent\_of},  \mathit{followed\_by}, \text{null} \}$ are more or less likely. Finally, the pair-wise features $\rho^{\text{pair}}_1$, $\rho^{\text{pair}}_2$, and $\rho^{\text{pair}}_3$ are then combined to a pair-wise output feature $\rho^{\text{pair}}_\text{out}$ that used to predict the $P_j^{\Psi}$ for all entity pairs.
\end{enumerate}
As a result, the entire component for relation classification is end-to-end trainable. This is a crucial difference of our \sysdgg/ over existing systems.

\vspace{-0.35cm}
\subsection{Grammar-based postprocessing (C4)}
\vspace{-0.15cm}

This component of our system converts hierarchical document structures $H_i$, $i = 1, \ldots, n$, consisting of the predicted entities $E$ and relations $R$, into a postprocessed document structure $H'$. Here, the aim is to ensure a \emph{valid}, tree-structured format that can later be used to generate different output formats such as hOCR \cite{breuelHOCRMicroformatOCR2007}. For this, we ensure that all entities form a tree structure w.r.t. their hierarchical relations of type $\Psi = \mathit{parent\_of}$ and are connected to a root entity with $c = \textsc{doc.\ root}$. We note that our postprocessing does not make any assumptions about the geometric overlap or the document layout. To this end, it is purely based on the document grammar and the predicted confidence scores $P^{\Psi}$. In particular, we apply our grammar-based postprocessing in sequential steps to address root entities ($g_{\text{rt}}$), illegal entities ($g_{\text{ilg}}$), and missing relations ($g_\text{mis}$) as follows:
\begin{itemize}
\item \textbf{Root entities} ($g_{\text{rt}}$): We append additional entities to build a basic skeleton for the document files. Specifically, we add root entities \textsc{doc.\ root}, \textsc{article} and \textsc{meta}. To enable full end-to-end training, we allow the prediction of these entities in the training process. 
\item \textbf{Illegal relations} ($g_{\text{ilg}}$): During training and inference of the relation classification, no restrictions are made on the possible combinations of entity pairs. This choice is made to allow for flexibility during end-to-end training (e.g., in the event that entities are not correctly predicted by the detection component of \sysdgg/). However, such flexibility can result in relations that violate our document grammar and where thus conflicts must be resolved. For example, we remove potential cycles so that the hierarchical document structure forms a tree structure. 
\item \textbf{Missing relations} ($g_\text{mis}$): Relations are added in order to ensure a valid tree structure so that each entity has a valid relation of the type $\Psi = \mathit{parent\_of}$. Specifically, every entity must have exactly one parent, except for the entity with $c = \textsc{doc. root}$, which has none. If an entity $E$ does not have a parent, we add a corresponding relation $(E_{\text{subj}}, E_{\text{obj}}, \Psi)$ with $E_{\text{obj}}=E$ and $\Psi = parent\_of$ and $E_{\text{subj}}$ based on the predicted confidence scores. 
\end{itemize}

\vspace{-0.35cm}
\subsection{hOCR conversion engine (C5)}
\vspace{-0.15cm}

In \sysdgg/, the final component is an hOCR conversion engine. It takes a postprocessed document structure $H'$ as input and then converts it into a hOCR file that is compatible with common open-source tools for document processing workflows. We extend the common hOCR format \cite{breuelHOCRMicroformatOCR2007} to additionally accommodate hierarchical structures.

\vspace{-0.35cm}
\subsection{Implementation details}
\vspace{-0.15cm}

Our \sysdgg/ system is based on the neural motifs architecture \cite{zellersNeuralMotifsScene2018} using the implementations from \cite{wu2019detectron2,khandelwalSegmentationgroundedSceneGraph2021}. However, we make non-trivial adaptations to accommodate our task of generating hierarchical document structures, as detailed in the following.

\textbf{Image preprocessing (component~C1):} Images are bilinearly resized so that their smallest side has at most size $\phi_s^\mathrm{max}$. If the longest side exceeds a predefined maximum size of $\phi_l^{\mathrm{max}}$ after this step, resizing is instead done so that the longest side length is $\phi_l^{\mathrm{max}}$. During training, the image size is varied for augmentation purposes. For this, $\phi_s^\mathrm{max}$ is chosen randomly from a set of different sizes. We set $\phi_l^{\mathrm{max}}$ at $600$, while $\phi_s^\mathrm{max}$ is randomly chosen from the range $[250, \ldots, 550]$ using increments of $50$. During training, images are randomly resized by applying the aforementioned resizing scheme.  For testing, we set $\phi_s^\mathrm{max} = 400$ and $\phi_l^{\mathrm{max}} = 600$. 

\textbf{Entity detection (component~C2):} The entity detection component is based on the Faster R-CNN architecture \cite{renFasterRCNNRealTime2015}.\footnote{We also experimented with a mask segmentation but could not measure a significant performance gain and, hence, discarded this subtask in our implementation.} We use a ResNet \cite{heDeepResidualLearning2016} backbone of depth $50$ in our system. Training of component~C2 uses the loss term $L_\text{C2}=L^\text{cls}_\text{RPN} + L^\text{loc}_\text{RPN} + L^\text{cls}_\text{E} + L^\text{loc}_\text{E}$. The losses $L^\text{loc}_\text{RPN}$ and $L^\text{cls}_\text{RPN}$ penalize localization and classification errors for the candidate regions generated by the region proposal network (RPN). Furthermore, the objective of correct localization and classification of predicted entities is formulated via the losses $L^\text{loc}_\text{E}$ and $L^\text{cls}_\text{E}$, respectively.

During training, up to $50$ entities are passed from the entity detection component (C2) to the component responsible for relation classification and entity refinement (C3).

\textbf{Relation classification (component~C3(i)):} The relation head uses a bidirectional LSTM with one recurrent layer, a hidden layer size of $512$, and a dropout of $0.2$. The relation context input features $\rho^{\text{rel\_in}}_1$ are extracted via the underlying Faster R-CNN architecture for the bounding box $B_j$ of each entity. Specifically, the convolutional neural network \cite{heDeepResidualLearning2016} of our Faster R-CNN architecture processes the input images in multiple sequential steps with decreasing spatial resolution. The output features of the Faster R-CNN are then fed into a feature pyramid network (FPN) \cite{Lin2017}, which produces multi-scale visual feature maps. Four multi-scale visual feature maps that correspond to the image region of $B_j$ are filtered by an alignment layer, concatenated, and passed through two fully-connected layers with ReLu activations \cite{nairRectifiedLinearUnits2010} to produce feature vectors of dimension $1024$. These feature vectors are then used as the relation context input feature $\rho^{\text{rel\_in}}_\text{vis}$. 

Following \cite{zellersNeuralMotifsScene2018}, we extract visual feature maps $\rho^{\text{pair}}_1$ for the union bounding box $B_\text{pair}$ of subject-object pairs $(E_{\text{subj}}, E_{\text{obj}})$. The extraction proceeds analogously to $\rho^{\text{rel\_in}}_\text{vis}$ but uses the region $B_\text{pair}$ to filter the multi-scale visual feature maps, resulting in pair-wise features $\rho^{\text{pair}}_1$. $\rho^{\text{pair}}_1$ and $\rho^{\text{pair}}_2$ are fed into fully-connected layers $W^{\text{pair}}_1$ and $W^{\text{pair}}_2$ with output dimensions of $4096$ and then combined using element-wise multiplication. The resulting feature vector is finally fed into a fully-connected layer $W^{\text{pair}}_\text{rel}$ with an output dimension equal to the number of relation types and added to the frequency bias term $\rho^{\text{pair}}_3$, resulting in the pair-wise output feature $\rho^{\text{pair}}_\text{out} = W^{\text{pair}}_\text{rel}((W^\text{pair}_1 \rho^{\text{pair}}_1)\circ (W^\text{pair}_2 \rho^{\text{pair}}_2))) + \rho^{\text{pair}}_3$. $\rho^{\text{pair}}_\text{out}$ is then used to predict the class probabilities for the relations. For training and evaluation, the ground-truth relation triples are matched to the candidate triples by calculating the IoU (see Sec.~\ref{sec:performance_metrics}) scores $S_{\text{subj}} = \text{IoU}(E_{\text{subj}}, E^{\text{GT}}_{\text{subj}})$ and $S_{\text{obj}} = \text{IoU}(E_{\text{obj}}, E^{\text{GT}}_{\text{obj}})$ of the subject and object entities in the relation, respectively. In accordance with our objective of uniquely matching all ground-truth relations, we allow only one candidate relation to be considered per ground-truth relation.

\textbf{Entity refinement (component~C3(ii)):} The refinement head is based on a bidirectional LSTM with one recurrent layer, a hidden layer size of $512$, and a dropout of $0.2$. The inputs to the LSTM are ordered according to their $x$-coordinates (center point) from left to right. The additional refinement context input features are computed as follows. The visual feature map $\rho^{\text{ref\_in}}_\text{vis}$ is computed analogously to $\rho^{\text{rel\_in}}_\text{vis}$. The category embedding $\rho^{\text{ref\_in}}_\text{cat}$ is computed by mapping the name of the semantic category directly onto the GloVe word embedding with an identical name \cite{penningtonGloveGlobalVectors2014}. For some categories that are encountered in our datasets, there is no direct match. For these categories, we use the following mapping: (\textsc{bibliography block} $\mapsto$ bibliography), (\textsc{text block} $\mapsto$ paragraph), (\textsc{figure caption} $\mapsto$ caption), (\textsc{figure graphic} $\mapsto$ graphic), (\textsc{page nr.} $\mapsto$ numbering), (\textsc{table caption} $\mapsto$ caption). The word embedding dimension is set to $200$ in our experiments. For the positional embedding $\rho^{\text{ref\_in}}_\text{pos}$, the bounding box width and $x$-coordinates are normalized with respect to the width of the full-sized image. Analogously, we normalize the box height and the $y$-coordinates with respect to the height of the full-sized image. 

Training of component~C3 uses the loss term $L_\text{C3}=L_\text{ref} + L_\text{rel}$, consisting of losses of relation classification, $L_\text{rel}$ and class refinement, $L_\text{ref}$.

\vspace{-0.35cm}
\section{Datasets}
\vspace{-0.15cm}

We compare our system using two different datasets.\footnote{A detailed overview is in our GitHub: \url{https://github.com/j-rausch/DSG}} Datasets that are suitable for evaluation must contain annotations for the full hierarchical document structure, and, hence, existing datasets are so far limited to scientific articles \cite{rauschDocParserHierarchicalDocument2021}. However, scientific articles follow a fairly similar structure. To this end, we also introduce a new dataset called \datasetEP/ containing real-world, offline magazines. This is beneficial for our evaluation as it provides a dataset with large heterogeneity in the presentation and thus complex document structures. The datasets are described in the following.

\vspace{-0.35cm}
\subsection{\datasetADtarget/}
\vspace{-0.15cm}
The existing dataset for training and evaluation, \datasetADtarget/, contains $362$ hand-annotated documents \cite{rauschDocParserHierarchicalDocument2021}. Previously, the dataset has only been used for training entity detection and not relation classification. To this end, we perform the following processing steps to allow end-to-end training. We first split any multi-page documents into separate single-page images. We then convert the dataset to a standardized format \cite{krishnaVisualGenomeConnecting2017} to allow processing with standardized benchmark libraries and document parsing codebases.

\vspace{-0.35cm}
\subsection{\datasetEP/}
\vspace{-0.15cm}

\datasetEP/ is a project that aims to digitize a wide range of historical and contemporary magazines for future generations \cite{wangerEPeriodicaPlattformFuer2018}. It comprises magazines from different source languages such as English, German, French, and Italian. Magazine pages often have a complex structure and follow little consistency in the formatting rules as compared to scientific literature.\footnote{Examples are on GitHub.} The entire \datasetEP/ contains over 8~million pages from over 400~journals. We manually annotated 542 documents comprising $11,446$ annotated entities. Specifically, we sampled a subset of document pages from journal issues of the past six decades. Moreover, the distribution of pages per journal is highly irregular, and, for this reason, we only consider five pages per issue of a given magazine in a given year. We then annotated entities and the relations between all entities to form hierarchical document structures. Details on the annotation procedure are provided below. We split the dataset into training, validation, and test sets of 270, 135, and 137 samples, respectively.

\textbf{Annotation procedure:} Our manual annotation followed a two-step process: (1)~entity annotation and (2)~relation annotation. In the entity annotation step, a bounding box is drawn around the entities on a page, and a semantic category is assigned to each entity. For the relation annotation step, we first annotate relations to define the reading order of a page by focusing on $\Psi = followed\_by$. If entities are nested, we additionally annotate relations that characterize nested structures given by $\Psi = parent\_of$ (e.g., (\textsc{figure}, \textsc{figure caption}, $parent\_of$)).

Specific considerations are made for the annotation of hierarchical structures in \datasetEP/. Unlike scientific documents, many magazine pages lack a standardized reading order (e.g., separate articles within a magazine can be read in arbitrary order). To model this, we designate two semantic categories for this purpose: \textsc{unordered group} and \textsc{ordered group}. An \textsc{unordered group} refers to parts that do not belong to the general document-level reading order (e.g., advertisements). An \textsc{ordered group} refers to parts that belong to the regular reading order (e.g., a single column with a separate article).

Further details and summary statistics are included in our GitHub.

\FloatBarrier

\vspace{-0.35cm}
\section{Experimental Setup}
\vspace{-0.25cm}

\subsection{Performance metrics}
\vspace{-0.15cm}
\label{sec:performance_metrics}

We separately evaluate the performance of our system for (i)~entity detection and (ii)~structure generation in which the hierarchical relations are considered. To this end, we adapt the benchmarking in related tasks from scene graph generation for our purpose of parsing document structures.

\textbf{Entity detection:} We follow common practice \cite{krishnaVisualGenomeConnecting2017,linMicrosoftCOCOCommon2014} in benchmarking entity detection. Specifically, we determine how many entities are correctly predicted out of all ground-truth entities. We compare the predicted entities $E_j = (c_j, B_j, P_j)$ with the ground-truth entities, consisting of the true category $\hat{c}_j$ and the true bounding box $\hat{B}_j$. Here, we measure the overlap between bounding boxes of the same category \cite{everinghamPASCALVisualObject2009}. Specifically, we calculate the so-called intersection-over-union (IoU) via
\begin{equation}
\text{IoU} = \frac{\mathit{area}(B_j \cap \hat{B}_j)}{\mathit{area}(B_j \cup \hat{B}_j)} .
\label{eq:iou_formula}
\end{equation} 
Predicted entities are considered a true positive if their IoU is higher than a pre-defined threshold. If more than one entity exceeds that threshold for the same ground-truth entity, the entity with the highest IoU is considered as a true positive. Any unmatched predicted entities and ground-truth entities are considered false positives and false negatives, respectively. We compare IoU thresholds of 0.5 and 0.75. All computations of the IoU are based on the API in \cite{linMicrosoftCOCOCommon2014}. 

We further calculate the average precision (AP) per semantic category $c_k \in \mathcal{C}$. The overall performance across all categories is given by the mean average precision (mAP) with (0: worst, 100: best).

\vspace{-0.35cm}
\subsection{Training and hyperparameter tuning}
\vspace{-0.15cm}

\textbf{Initialization:} We initialize our systems with two pre-training steps. First, the systems are initialized with weights of a Faster R-CNN architecture trained on the COCO dataset \cite{linMicrosoftCOCOCommon2014} with copy-paste augmentation \cite{ghiasiSimpleCopyPasteStrong2021}. Second, since the COCO dataset does not contain documents, we then proceed to pre-train all systems on \datasetADweak/ \cite{rauschDocParserHierarchicalDocument2021}. Like \datasetADtarget/, this dataset has been generated for scientific articles, but it was only annotated with a weak supervision mechanism. This allows for better preparation of the document parsing tasks. The resulting system weights are then used as a starting point for our experiments.

\textbf{Training:} We first sample $128$ from all possible entity-entity pairs per training iteration to serve as input to the relation classification of the relation head. This reduces the computational complexity of the training and allows us to sample a more balanced set of positive and negative samples, since the majority of entity-entity pairs correspond to relations of type $\Psi =\text{null}$. Unlike common training procedures from scene graph generation, we do not apply geometric constraints on candidate pairs $(E_{\text{subj}}, E_{\text{obj}})$. Concretely, this means that entity-entity pairs with no geometric overlap are considered for relation prediction. This is especially important for relations of type $\Psi =   \mathit{followed\_by}$, where the bounding boxes $(B_{\text{subj}}$ and $B_{\text{obj}})$ of $(E_{\text{subj}}$ and $E_{\text{obj}})$ do not intersect. In order to allow for training of the refinement and relation head throughout the whole training procedure, we append any missing ground-truth entities to the set of entities that are passed to the refinement head. This is to avoid cases where no entities or only erroneous entities are detected and where thus no positive learning samples can be provided. This happens, for example, at the beginning of the training procedure.

We train our \sysdgg/ system end-to-end via a joint objective consisting of (i)~the entity detection component based on the Faster R-CNN component and (ii)~the relation classification and entity refinement component. The losses $L_\text{C2}$ and $L_\text{C3}$ for both components are combined to train our \sysdgg/ system. Our system is trained for up to $200,000$ iterations with a batch size of $4$ and a learning rate of $0.001$. We apply early stopping based on the performance for entity detection on the validation set.

\textbf{Computational performance:} We measure the computational performance of our system on a machine with a single NVIDIA Titan Xp GPU with a memory size of 12GB. Here, the average time to process a sample, as measured on the validation set with a batch size of $1$, is $\sim0.1616$ seconds. Integrating our grammar-based postprocessing comes with only a small overhead and results in an average joint processing time of $\sim0.1776$ seconds for the same computational setup.

\vspace{-0.35cm}
\subsection{Baselines}
\vspace{-0.15cm}

We benchmark our proposed \sysdgg/ against state-of-the-art systems for document structure parsing.
\begin{itemize} 
\item \textbf{DocParser} \cite{rauschDocParserHierarchicalDocument2021}:  We reimplement DocParser \cite{rauschDocParserHierarchicalDocument2021}, which is a state-of-the-art system for hierarchical structure parsing. To the best of our knowledge, this is the only suitable baseline that can parse entire document structures (see Sec.~\ref{sec:related_work}). To ensure comparability between DocParser and our system, we use the Faster R-CNN architecture with a ResNet \cite{heDeepResidualLearning2016} backbone of depth $50$ for entity detection. Of note, DocParser uses heuristics for the relation classification and is thus \emph{not} end-to-end trainable. Therefore, we evaluate DocParser using the original heuristics in \cite{rauschDocParserHierarchicalDocument2021} for relation classification. We extend the heuristics to accommodate the additional root entity \textsc{article} (i.e., map it onto the entity category \textsc{document}). 
\end{itemize}

\noindent
We further compare different variants of \sysdgg/ that act as ablation studies to demonstrate the importance of end-to-end training. Thereby, we can evaluate the importance of end-to-end vs. 2-stage training. 
\begin{itemize} 
\item \textbf{\sysdgg/ 2-stage (C2 frozen):} In the first stage, component C2 of \sysdgg/ is trained exclusively with respect to the correct prediction of entities. In the second stage, component C3 is added to \sysdgg/ training. However, the weights of component C2 are frozen during the second stage. Hence, the loss to update C3 is only based on the predictions of the relation head and refinement head.
\item
\textbf{\sysdgg/ 2-stage (C2 unfrozen):} In the first stage, component C2 of \sysdgg/ is trained. Here, a loss is used which only learns against the correct prediction of entities. In the second stage, we allow parameter weight updates to both C2 and C3 of the system. Here, a loss is used which only learns against the predictions of the relation head and refinement head of component C3, but not against the prediction of entities by component C2. 
\item \textbf{\sysdgg/ end-to-end (w/o postprocessing)}: This is our \sysdgg/ system from above that is trained in an end-to-end fashion but without the postproecessing from component C4. 
\item \textbf{\sysdgg/ end-to-end (w/ postprocessing)}: This is our \sysdgg/ system from above that is trained in an end-to-end fashion. 
\end{itemize}

\noindent
\textbf{Training procedure for baselines:} 
We use identical hyperparameter settings for training the entity detection component that are used for the baseline systems DocParser, \sysdgg/ 2-stage, and \sysdgg/ end-to-end. All systems are trained for up to $100,000$ iterations with a learning rate of $0.01$ and a batch size of $8$. This training only includes the prediction of document entities through component C2 and uses the loss term $L_\text{C2}$. For DocParser, we apply early stopping based on the mAP score for an IoU threshold of $0.5$ on the validation set. 

After training the component C2 for entity detection, we continue with the second training stage of \sysdgg/ 2-stage (C2 frozen) and \sysdgg/ 2-stage (C2  unfrozen), where the systems are trained with a joint objective for relation classification and entity refinement using the loss term $L_\text{C3}$. We perform the second stage with a learning rate of $0.001$ and a batch size of $8$ for up to $100,000$ iterations. Unlike \sysdgg/, the systems in the ablation study do not use the additional loss $L_\text{C2}$ for entity detection via component C2 in this stage. We apply early stopping based on the mAP score for an IoU threshold of $0.5$ on the validation set in the second stage.

\textbf{Structure generation:} To evaluate how well the hierarchical structure is generated, we perform an evaluation of triplets for the relations. Specifically, we measure exact matches of the predicted relations $(E_{\text{subj}}, E_{\text{obj}}, \Psi)$ against the ground-truth observations and report the corresponding F1 score. The F1 score is the harmonic average of precision and recall for predicting these triples, i.e., $F_1 = 2\frac{\text{precision}\cdot\text{recall}}{\text{precision}+\text{recall}}$ with (0: worst, 1: best). For this, we first determine all matches of bounding boxes of entities. Subsequently, a relation is considered a match if the predicted relation type \emph{and} both bounding boxes match with a ground-truth triple.

Our above performance measures are fairly strict when compared with evaluations in conventional scene graph generation (e.g., \cite{zellersNeuralMotifsScene2018,khandelwalSegmentationgroundedSceneGraph2021}). Recall that we consider a match if and only if the relation type \emph{and} both bounding boxes match with a \emph{single} ground-truth triplet. In contrast, conventional scene graph generation uses a relaxed definition where a predicted relation triple $(E^p_1, E^p_2, \phi^p)$ is considered a match with a ground-truth triple $(E^g_1, E^g_2, \phi^p)$ if $\phi^p = \phi^g$ and if an IoU overlap is found between the ground-truth entities and both predicted entities. However, this definition allows for that more than one predicted entity could be matched with a ground-truth entity during evaluation of relation prediction. In our evaluation, we apply a more strict performance that considers at most one \emph{unique} match with ground-truth entity, following the same procedure as during entity detection evaluation.

Our choice of performance metrics is important to effectively differentiate between closely nested entities. This is relevant, for example, to distinguish between a figure that is wrapped around a subfigure. Using simple IoU matching, a predicted entity could be matched with either of the two ground-truth candidate entities (i.e., the figure and the subfigure). However, to reconstruct the hierarchical document structure, it is crucial to correctly determine the exact hierarchical relations among entities to arrive at a unique and valid tree structure.

\vspace{-0.35cm}
\section{Results}
\vspace{-0.25cm}

\subsection{Numerical Experiments}
\vspace{-0.15cm}

The objective of our experiments is to confirm the effectiveness of our \sysdgg/ in generating the hierarchical document structures. Hence, we proceed two-fold: (1)~We first measure the performance in entity detection, and (2)~we measure the performance correctly generating hierarchical structures. 

\textbf{Entity detection:} We report the performance for entity detection in Table~\ref{tab:arxvivdocs_detection_res_test1} (\datasetADtarget/) and Table~ \ref{tab:eperiodica_detection_res_test1} (\datasetEP/). Here, we report the results for our \sysdgg/ (last rows). We further state the results for DocParser \cite{rauschDocParserHierarchicalDocument2021}, which is a state-of-the-art baseline and which is the only existing system for our task. We further compare different variants of our systems to assess the importance of end-to-end vs. 2-stage training. 

We make the following observations. (1)~The performance in entity detection is generally better for \datasetADtarget/ (scientific articles) than for \datasetEP/ (magazines). This demonstrates that our new dataset is a challenging, real-world setting for evaluation. In particular, the performance difference can be explained by that the format of magazines is characterized by a fairly large variability compared to scientific articles. (2)~Our \sysdgg/ with end-to-end training consistently performs best. In particular, it outperforms the state-of-the-art DocParser from \cite{rauschDocParserHierarchicalDocument2021}. For example, for an IoU threshold of 0.5, the mAP of our end-to-end \sysdgg/ is 1.91 percentage points better than DocParser for \datasetADtarget/, and it is 7.03 percentage points better for \datasetEP/. This translates into a relative improvement of $2.46\%$ and $12.71\%$, respectively. (3)~The larger relative performance gain for \datasetEP/ than for \datasetADtarget/ is likely due to the fact that our system can directly leverage annotated relations and learn against them, whereas DocParser is limited to simple heuristics. (4)~Our \sysdgg/ using end-to-end training outperforms the 2-stage training approach. Hence, one of the reasons for the strong performance of our approach is the joint learning procedure, in which components C2 and C3 allow for system-wide parameter updates. In sum, the results demonstrate the effectiveness of our \sysdgg/.

\begin{table}[tbh!]
\footnotesize
\begin{tabular}{l cc cc}
\toprule
\textbf{Semantic category} &  \multicolumn{2}{c}{\textbf{DocParser \cite{rauschDocParserHierarchicalDocument2021}}} & \multicolumn{2}{c}{\textbf{\sysdgg/ (ours)}}\\
\cmidrule(lr){2-3}  \cmidrule(lr){4-5}
& \textbf{IoU=0.5} &  \textbf{IoU=0.75} & \textbf{IoU=0.5} &  \textbf{IoU=0.75} \\
\midrule
mAP  &  55.32  &  35.33  &  \textbf{62.35}  &  \textbf{41.26} \\
\midrule
 $\textsc{article}$  &  66.51  &  48.69  &  \textbf{75.85}  &  \textbf{59.43}  \\
 $\textsc{author}$  &  41.32  &  16.92  &  \textbf{47.40}  &  \textbf{20.40}  \\
 $\textsc{backgr.\_fig.}$  &  53.69  &  34.66  &  \textbf{69.45}  &  \textbf{47.72}  \\
 $\textsc{column}$  &  45.36  &  \textbf{23.44}  &  \textbf{70.00}  &  18.87  \\
 $\textsc{textblock}$  &  78.03  &  66.86  &  \textbf{80.28}  &  \textbf{68.21}  \\
 $\textsc{doc.\_root}$  &  \textbf{99.01}  &  \textbf{99.01}  &  \textbf{99.01}  &  \textbf{99.01}  \\
 $\textsc{figure}$  &  42.10  &  19.08  &  \textbf{56.64}  &  \textbf{36.75}  \\
 $\textsc{fig.\_caption}$  &  32.58  &  23.15  &  \textbf{41.10}  &  \textbf{31.45}  \\
 $\textsc{fig.\_graphic}$  &  66.29  &  50.10  &  \textbf{76.55}  &  \textbf{58.32}  \\
 $\textsc{footer}$  &  42.59  &  8.51  &  \textbf{56.99}  &  \textbf{18.07}  \\
 $\textsc{footnote}$  &  57.97  &  45.83  &  \textbf{66.86}  &  \textbf{59.12}  \\
 $\textsc{header}$  &  \textbf{44.83}  &  \textbf{15.87}  &  32.24  &  13.36  \\
 $\textsc{heading}$  &  57.61  &  26.07  &  \textbf{65.17}  &  \textbf{35.42}  \\
 $\textsc{item}$  &  46.76  &  30.62  &  \textbf{56.04}  &  \textbf{37.01}  \\
 $\textsc{itemize}$  &  53.72  &  40.87  &  \textbf{69.73}  &  \textbf{56.40}  \\
 $\textsc{meta}$  &  83.97  &  83.97  &  \textbf{87.23}  &  \textbf{87.23}  \\
 $\textsc{orderedgroup}$  &  65.51  &  42.02  &  \textbf{70.38}  &  \textbf{49.47}  \\
 $\textsc{page\_nr.}$  &  62.93  &  1.80  &  \textbf{68.21}  &  \textbf{7.22}  \\
 $\textsc{row}$  &  49.75  &  17.00  &  \textbf{52.12}  &  \textbf{23.28}  \\
 $\textsc{table}$  &  52.87  &  39.38  &  \textbf{61.26}  &  \textbf{48.61}  \\
 $\textsc{table\_of\_cont.}$  &  \textbf{51.90}  &  \textbf{21.99}  &  47.18  &  20.20  \\
 $\textsc{tabular}$  &  52.58  &  \textbf{37.84}  &  \textbf{57.54}  &  31.28  \\
 $\textsc{unordered\_group}$  &  24.52  &  18.99  &  \textbf{26.79}  &  \textbf{22.20}  \\
\bottomrule
\end{tabular}
\caption{Performance of entity detection per category on \datasetEP/. Reported: average precision (AP) per semantic category. Compared: DocParser and \sysdgg/ end-to-end.}
\vspace{-0.35cm}
\label{tab:eperiodica_detection_res_test_per_category}
\end{table}

Table~\ref{tab:eperiodica_detection_res_test_per_category} provides a breakdown by different semantic categories on the E-Periodica dataset. Evidently, our \sysdgg/ is consistently better for the vast majority of semantic categories.\footnote{A detailed breakdown by different semantic categories is provided in our GitHub at \url{https://github.com/j-rausch/DSG}} For example, for an IoU threshold of 0.5, it achieves an improvement by $9.34$ percentage points for the \textsc{article} category. Entities of this category are important during relation classification, because they reflect the high-level segmentation of document pages and, thus, are used in a large number of hierarchical relations. Evidently, the end-to-end training objective of \sysdgg/ that incorporates relation-level losses provides the system with useful supervision signals for this category.  DocParser scores slightly better than our system for the \textsc{header} category at an IoU threshold of 0.5. We hypothesize that this could be due to the fact that \textsc{header} entities only account for $1.38\%$ of all entities and are less relevant for parsing the document structure, since they are often not part of the reading order in magazine articles. As such, \sysdgg/ is less incentivized to optimize for this entity category through its end-to-end training objective.

\begin{table}[h!]
\centering
\footnotesize
\begin{tabular}{llcc}
\toprule
\textbf{System} & \textbf{Variant} & \textbf{IoU=0.5} &  \textbf{IoU=0.75} \\
\midrule
DocParser \cite{rauschDocParserHierarchicalDocument2021} & -- & 77.70 & 58.62 \\
\midrule 
\sysdgg/ 2-stage (ours) & C2 frozen & 71.03 & 50.48 \\
\sysdgg/ 2-stage (ours) & C2 unfrozen & 77.48 & 57.40 \\
\midrule
\sysdgg/ end-to-end (ours) & w/o postproc. & \textbf{79.61} & \textbf{58.58} \\
\sysdgg/ end-to-end (ours) & w/ postproc. & \textbf{79.61} & \textbf{58.58} \\
\bottomrule
\end{tabular}
\caption{Performance (mAP) of entity detection on \datasetADtarget/.}
\vspace{-0.5cm}
\label{tab:arxvivdocs_detection_res_test1}
\end{table}

\begin{table}[h!]
\centering
\footnotesize
\begin{tabular}{llrr}
\toprule
\textbf{System} & \textbf{Variant} & \textbf{IoU=0.5} &  \textbf{IoU=0.75} \\
\midrule
DocParser &  -- & 55.32 & 35.33 \\
\sysdgg/ 2-stage (ours) & C2 frozen & 51.51 & 10.84 \\
\sysdgg/ 2stage (ours) & C2 unfrozen & 54.41 & 36.22 \\
\midrule
\sysdgg/ end-to-end (ours) & w/o postproc. & \textbf{62.35} & \textbf{41.26} \\
\sysdgg/ end-to-end (ours) & w/ postproc. & \underline{62.18} & \underline{40.90} \\
\bottomrule
\end{tabular}

    \caption{Performance of entity detection on \datasetEP/.}
\vspace{-0.25cm}
    \label{tab:eperiodica_detection_res_test1}
\end{table}

\textbf{Structure generation:} We now evaluate the accuracy with which the hierarchical relations are correctly generated. 
For this, we again report the performance for both datasets, namely, \datasetADtarget/ (Table~\ref{tab:arxvivdocs_structure_res_test1}) and \datasetEP/ (Table~\ref{tab:eperiodica_structure_res_test1}). 

We make the following observations. (1)~We again measure an overall better performance for \datasetADtarget/ than for \datasetEP/. This is expected due to the complex format of magazine articles. (2)~We find that our \sysdgg/ performs best. In particular, it outperforms the state-of-the-art DocParser \cite{rauschDocParserHierarchicalDocument2021} from the literature by a clear margin. Our system improves over the F1 from DocParser by 7.63\% (\datasetADtarget/) and 183.44\% (\datasetEP/). (3)~We again find a larger improvements for \datasetEP/ than for \datasetADtarget/. This can be explained by that our system can directly leverage annotated relations and learn against them, whereas DocParser is limited to simple heuristics. (4)~Our \sysdgg/ benefits from end-to-end training. As can be seen in our ablation studies, end-to-end training outperforms 2-stage training. (5)~We observe a slight drop in F1 scores after applying postprocessing the hierarchical structures $H$ produced by \sysdgg/. A reason for this lies in our strict evaluation procedure, paired with the motivation of producing valid tree structures as a result of our postprocessing. To illustrate this, let us consider an entity that is missed by our system. If this node would be normally be positioned as an intermediate node in the document, our postprocessing could connect its successor and predecessor entities to form a valid document structure. This would, however, have a negative effect on overall performance, but facilitates our aim of generating valid tree structures.

The above evaluation has also an important implication. DocParser builds on heuristics that were specifically tailored to scientific articles in the \datasetADtarget/ dataset. For this reason, DocParser is \emph{not} directly effective for other datasets such as \datasetEP/ without manual re-engineering. 

We remind that we enforce a strict evaluation in which the complete tuple including both entities must be correct. Hence, our structure parsing task relies on the accurate identification of \emph{every} entity and the relation type in a given triplet. Because of this, high F1 scores require a high detection accuracy in the entity recognition. Nevertheless, the performance of our system is highly effective in practice where the aim is to recover the overall document structure.

\textbf{Qualitative assessment :} We performed a qualitative assessment (see our GitHub at \url{https://github.com/j-rausch/DSG}). Thereby, we demonstrate that we generate meaningful and effective document structures in practice.

\begin{table}[h!]
\footnotesize
\begin{tabular}{ll ccc}
\toprule
\textbf{System} & \textbf{Variant} & \textbf{Precision} &  \textbf{Recall} & \textbf{F1} \\
\midrule
DocParser \cite{rauschDocParserHierarchicalDocument2021} & -- & 0.6646 & \textbf{0.7687} & 0.7054 \\
\midrule
\sysdgg/ 2-stage (ours) & C2 frozen & \underline{0.7689} & 0.7042 & 0.7223 \\
\sysdgg/ 2-stage (ours) & C2 unfrozen & 0.7378 & 0.7560 & \underline{0.7378} \\
\midrule
\sysdgg/ end-to-end (ours) & w/o postproc. & \textbf{0.7709} & \underline{0.7649} & \textbf{0.7592} \\
\sysdgg/ end-to-end (ours) & w/ postproc.  & 0.6959 & 0.7590 & 0.7185 \\
\bottomrule
\end{tabular}
    \caption{Performance of structure parsing on \datasetADtarget/.}
\vspace{-0.5cm}
    \label{tab:arxvivdocs_structure_res_test1}
\end{table}

\begin{table}[h!]
\footnotesize
\begin{tabular}{ll ccc}
\toprule
\textbf{System} & \textbf{Variant} & \textbf{Precision} &  \textbf{Recall} & \textbf{F1} \\
\midrule
DocParser \cite{rauschDocParserHierarchicalDocument2021} & -- & 0.1725 & 0.2319 & 0.1884 \\
\midrule
\sysdgg/ 2-stage (ours) & C2 frozen & 0.3901 & 0.5083 & 0.4232 \\
\sysdgg/ 2-stage (ours) & C2 unfrozen & 0.4589 & \underline{0.5276} & 0.4740 \\
\midrule
\sysdgg/ end-to-end (ours) & w/o postproc. & \underline{0.5545} & \textbf{0.5528} & \textbf{0.5340} \\
\sysdgg/ end-to-end (ours) & w/ postproc. & \textbf{0.5701} & 0.5197 & \underline{0.5308} \\
\bottomrule
\end{tabular}
    \caption{Performance of structure parsing on \datasetEP/.}
    \vspace{-0.5cm}
    \label{tab:eperiodica_structure_res_test1}
\end{table}

\vspace{-0.35cm}
\section{Discussion}
\vspace{-0.15cm}

\textbf{Novel system:} Our system is relevant for several downstream tasks for which document renderings (e.g., PDF files, scans) must be mapped onto a parseable format. Examples are \cite{cheQueryOptimizationXML2006,cafarellaWebTablesExploringPower2008,wilkinsonEffectiveRetrievalStructured1994,liIndexingQueryingXML2001,manabeExtractingLogicalHierarchical2015,wuFonduerKnowledgeBase2018}. Recent works (e.g., \cite{appalarajuDocFormerEndtoEndTransformer2021a,huangLayoutLMv3PretrainingDocument2022a}) have introduced transformer-based systems for large-scale pre-training on document data but for other tasks such as entity detection and thus \emph{without} extracting hierarchical structures. Hence, our system is orthogonal to such works and makes an important, non-trivial contribution. Importantly, the main advantage of our \sysdgg/ is that it can generate complete hierarchical document structures through end-to-end training.

\textbf{Comparison to OCR systems:} Prior research (e.g., \cite{liEndtoEndOCRText2020a}) has repeatedly demonstrated the challenges in existing OCR systems. OCR systems are typically not designed for generating hierarchical document structures but primarily for inferring textual contents from document renderings. As a result, OCR systems generally struggle with recognizing fine-grained structures such as subfigures and their ordering (see our qualitative analysis above). Our system alleviates these challenges and is thus specifically designed to accurately generate hierarchical document structures with high granularity to enable downstream tasks. To this end, we opted for relatively strict performance metrics to ensure that fine-grained structures are recognized correctly.

\textbf{Practical strengths:} A key strength of our system is that it is end-to-end trainable. This allows our system to take full advantage of existing training data, including information on the hierarchical relations that captures the sequence and nested structure within documents. In contrast, prior systems \cite{rauschDocParserHierarchicalDocument2021} are \emph{not} end-to-end trainable but infer relations through heuristics, thereby essentially ignoring the corresponding information in the training data. As a result, our system reduces the cost of annotating hierarchical document structures by a significant extent. In sum, our system fulfills a key demand in practice where the generation of document structures is often subject to scarce data and where systems should be customizable in a flexible manner.

\textbf{Novel dataset:} We contribute a novel, large-scale dataset based on magazines for generating hierarchical document structures. In particular, our dataset provides a challenging real-world setting for evaluation due to the large heterogeneity in the layout of magazines. Key to our dataset is its large granularity of the annotations in terms of both fine-grained entities and the relations between them. This is different from other datasets, which are typically coarse \cite{zhongPubLayNetLargestDataset2019a} and without hierarchical information \cite{liDocBankBenchmarkDataset2020b}.

\textbf{Conclusion:} In this paper, we introduced \emph{\syslong/} (\sysdgg/), a novel system for parsing hierarchical document structures that is end-to-end trainable. We show that our system outperforms state-of-the-art systems. By being end-to-end trainable, our \sysdgg/ is of direct value in practice in that it can be adapted to new documents in a straightforward manner without the need for manual re-engineering.

\FloatBarrier

\bibliographystyle{IEEEtran}
\vspace{-0.35cm}
\bibliography{references}%

\begin{appendices}

\section{Related Work}
Table~\ref{tab:related_work_overview} shows an overview of key systems for document structure parsing. 

\begin{table}[tbh!]
\footnotesize
\centering
\begin{tabular}{lccc}
\toprule
\textbf{System} & \makecell{\textbf{Document}\\\textbf{entities}} & \makecell{\textbf{Hierarchical}\\\textbf{structures}} & \makecell{\textbf{End-to-end}\\\textbf{training}} \\
\midrule
PubLayNet \cite{zhongPubLayNetLargestDataset2019a} & \cmark & \xmark & \cmark \\
DocFormer \cite{appalarajuDocFormerEndtoEndTransformer2021a} &\cmark & \xmark  & \cmark \\
UniDoc \cite{guUniDocUnifiedPretraining2021} &  \cmark & \xmark & \cmark\\
DiT \cite{liDiTSelfsupervisedPretraining2022a} &  \cmark & \xmark & \cmark\\
HRDoc \cite{maHRDocDatasetBaseline2023} & only line-level & \cmark & \cmark \\
TableBank \cite{liTableBankBenchmarkDataset2020a} &  only tables  & \cmark & \cmark \\ %
PubTables \cite{smockPubTables1MComprehensiveTable2022} & only tables & \cmark & \xmark\\ %
DocParser \cite{rauschDocParserHierarchicalDocument2021} & \cmark & \cmark & \xmark \\
\midrule
\sysdgg/ (\emph{ours}) & \cmark & \cmark & \cmark \\
\bottomrule
\end{tabular}
    \caption{Overview of key systems for document structure parsing from document renderings.}
    \label{tab:related_work_overview}
\end{table}

\section{Our \sysdgg/ System}

\subsection{Grammar-based postprocessing (C4)}

\noindent
\textbf{$g_{\text{ilg}}$ Illegal relations:}
\begin{itemize}
    \item We enforce that the root entity of type \textsc{doc. root} can only be part of relations in which it is a parent entity.  
    \item All relations are anti-symmetric. For this, we ensure that no two relations with $\Psi \in \mathit{\{parent\_of, followed\_by\}}$ exist that result in a symmetric relation. We resolve such cases by deleting the conflicting relations with lower confidence.
    \item No two relations with the same $\Psi \in \mathit{\{followed\_by, parent\_of\}}$ end in the same entity. 
    \item We ensure that entities can only be followed by at most one other entity.
    \item We ensure that entities of the \textsc{unordered group} category are not part of any sequential relations.
    \item We remove any cycles that might be formed by the graph formed by the predicted relations.
    \item Sequential relations can only exist between sibling entities belonging to the same parent entity.
\end{itemize}

\noindent
\textbf{$g_\text{mis}$ Missing relations:} 
If, after performing the previous postprocessing steps, an entity does not have a parent entity, we inspect the confidence scores of all relations $(E_\text{cand},E_\text{missing\_parent},\Psi)$ with $\Psi = \mathit{parent\_of}$ and candidate parent entities $E_\text{cand}$. We retrieve the relation with the highest confidence score, even if this score would otherwise not be sufficiently high to determine a relation with $\Psi = \mathit{parent\_of}$ and ensure that the resulting relation adheres to $g_{\text{ilg}}$.

\subsection{hOCR conversion engine (C5)}

The hOCR format \cite{breuelHOCRMicroformatOCR2007} encodes information using extensible markup language (XML) and is built upon hypertext markup language (HTML). To ensure compatibility with standard hOCR tools while still accommodating the hierarchical structure from \sysdgg/, we add a new \sysdgg/-specific XML node \texttt{<div>} into the standard hOCR XML nodes. Our new node does not have a hOCR-specific class attribute. As a consequence, third-party tools are still able to process our output as valid hOCR, since only hOCR-specific class attributes are considered by default. In plain words, the hierarchy and sequential ordering are preserved, while additional information from our \sysdgg/, such as the extended set of semantic categories, is ignored.

We use specific hOCR elements to convert the postprocessed hierarchical structure $H'$ into an hOCR file. We use a mapping to create the hOCR XML nodes from \sysdgg/ entities. Formally, we specify a mapping $\omega : c \rightarrow \eta$ of the \sysdgg/ entity categories $c$ to the hOCR element classes, denoted by $\eta$. We provide a list of $(c, \eta)$ tuples in Table~\ref{tab:dgg_to_hocr_mapping}. The hOCR format does not have elements that match the semantic categories of \textsc{meta} and \textsc{article} in \sysdgg/. In order to deal with both semantic categories, we keep the \sysdgg/-specific XML nodes to preserve the underlying structural information.

\begin{table*}[h!]
\footnotesize
\centering
\begin{tabular}{ll|ll}
\toprule
 \sysdgg/ semantic cat. & hOCR class & \sysdgg/ semantic cat & hOCR class \\
 \midrule
\textsc{doc. root} & \texttt{ocr\_page}  & \textsc{item} &  \texttt{ocr\_carea} \\
\textsc{meta} &  \texttt{None}  & \textsc{itemize} &  \texttt{ocr\_float} \\
\textsc{author} &  \texttt{ocr\_author}  & \textsc{ordered group} &  \texttt{ocr\_carea} \\
\textsc{background fig.} &  \texttt{ocr\_float}  & \textsc{page nr.} &  \texttt{ocr\_pageno} \\
\textsc{text block} &  \texttt{ocrx\_block}  & \textsc{table} &  \texttt{ocr\_table} \\
\textsc{figure} &  \texttt{ocr\_float}  & \textsc{tabular} &  \texttt{ocr\_table} \\
\textsc{figure graphic} &  \texttt{ocr\_photo}  & \textsc{table of contents} &  \texttt{ocr\_table} \\
\textsc{figure caption} &  \texttt{ocr\_caption}  & \textsc{unordered group} &  \texttt{ocr\_float} \\
\textsc{footer} &  \texttt{ocr\_footer}  & \textsc{article} &  \texttt{None} \\
\textsc{footnote} &  \texttt{ocr\_footer}  & \textsc{column} &  \texttt{ocr\_carea} \\
\textsc{header} &  \texttt{ocr\_header}  & \textsc{row} &  \texttt{ocr\_carea} \\
\textsc{heading} &  \texttt{ocr\_header} &&\\
\bottomrule
\end{tabular}
    \caption{Mapping between semantic categories in \sysdgg/ and hOCR classes.}
    \label{tab:dgg_to_hocr_mapping}
\end{table*}

The conversion process consists of three steps to convert the postprocessed hierarchical structures $H'$ into hOCR files:
\begin{enumerate}
\item \textbf{Initialization ($s_1$):} We initialize the hOCR file with the hierarchical structure $H'_\mathit{parent\_of}$ that only considers relations of type $\Psi = \mathit{parent\_of}$. For this, hOCR and additional \sysdgg/ XML nodes are initialized according to $H'_\mathit{parent\_of}$ and the category mapping $\omega$.
\item \textbf{Order of children ($s_2$):} We ensure a correct ordering of the children. Formally, we ensure that, if any $\Psi = \mathit{followed\_by}$ relation exists between two entities in $H'$, the corresponding XML nodes follow that order.
\item \textbf{OCR enrichment ($s_3$):} We additionally enrich the hOCR files with the textual contents $T$ of the documents. 
First, we ensure that words are only appended to leaf node entities $E_\text{leaf}$ in the document structure. Second, words are assigned to the entity $E_\text{leaf}$ with the highest intersection-over-union score (see Sec.~\ref{sec:performance_metrics}) between entity bounding box $B_\text{leaf}$ and word bounding box $B_\text{word}$.
\end{enumerate}

\noindent
A key feature of the hOCR format is the ability to perform structure-based XPath queries on document files \cite{breuelHOCRMicroformatOCR2007}. To facilitate such applications in practice, we extend the XPath queries to account for the hierarchical structures in our hOCR files generated by \sysdgg/. As a result, we allow that hOCR files can be searched for specific \sysdgg/ entities and relations using XPath queries. We provide XPath queries for three different types of queries: 
\begin{enumerate}
\item \textbf{Node name search ($q_1$):} This is a simple search by node name that returns all XML nodes that match the desired node name.
\item \textbf{Absolute path search ($q_2$):} We offer the ability to navigate documents using absolute paths. An absolute path starts with a $/$ symbol and then describes the path to the desired node starting from the root node of the XML file. Here, the different names of the nodes along this path are concatenated using $/$ symbols.
\item \textbf{Relative path search ($q_3$):} Relative paths can be used to retrieve a desired node. For this, two wildcard elements are used to match any potential node or attribute. The starting symbol $//$ indicates a relative path search. The relative path search then returns any node that has a path that could potentially match the query by using two wildcard elements $*$ to match any node or path and $@*$ to match any node attribute. 
\end{enumerate}

\section{Datasets}

\begin{table}[h!]
\footnotesize
\centering
\begin{tabular}{lllcc}
\toprule
         Dataset  & Document type & \#Docs & \#Categories & Source\\
\midrule
     \datasetADtarget/ & Scientific articles   & $362$ & $21$ & \cite{rauschDocParserHierarchicalDocument2021}  \\
     \datasetEP/ & Magazines  & $542$ & $22$ & \emph{Ours}  \\
\bottomrule
\end{tabular}
    \caption{Overview of existing datasets for our task.}
    \label{tab:datasets_overview}
\end{table}
An overview of the two datasets used in our experiments is given in Table~\ref{tab:datasets_overview}.
\subsection{\datasetEP/}

\begin{figure}[tbh!]
\centering
\begin{subfigure}[t]{0.49\linewidth}
\centering
\includegraphics[width=\linewidth,frame]{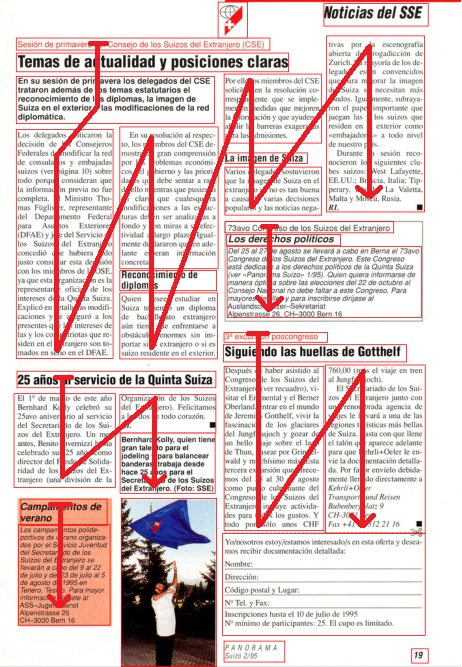}
\caption{Example of a complex reading order.}
\label{fig:EP_example2_reading_order}
\end{subfigure}
\hfill
\begin{subfigure}[t]{0.49\linewidth}
\centering
\includegraphics[width=\linewidth,frame]{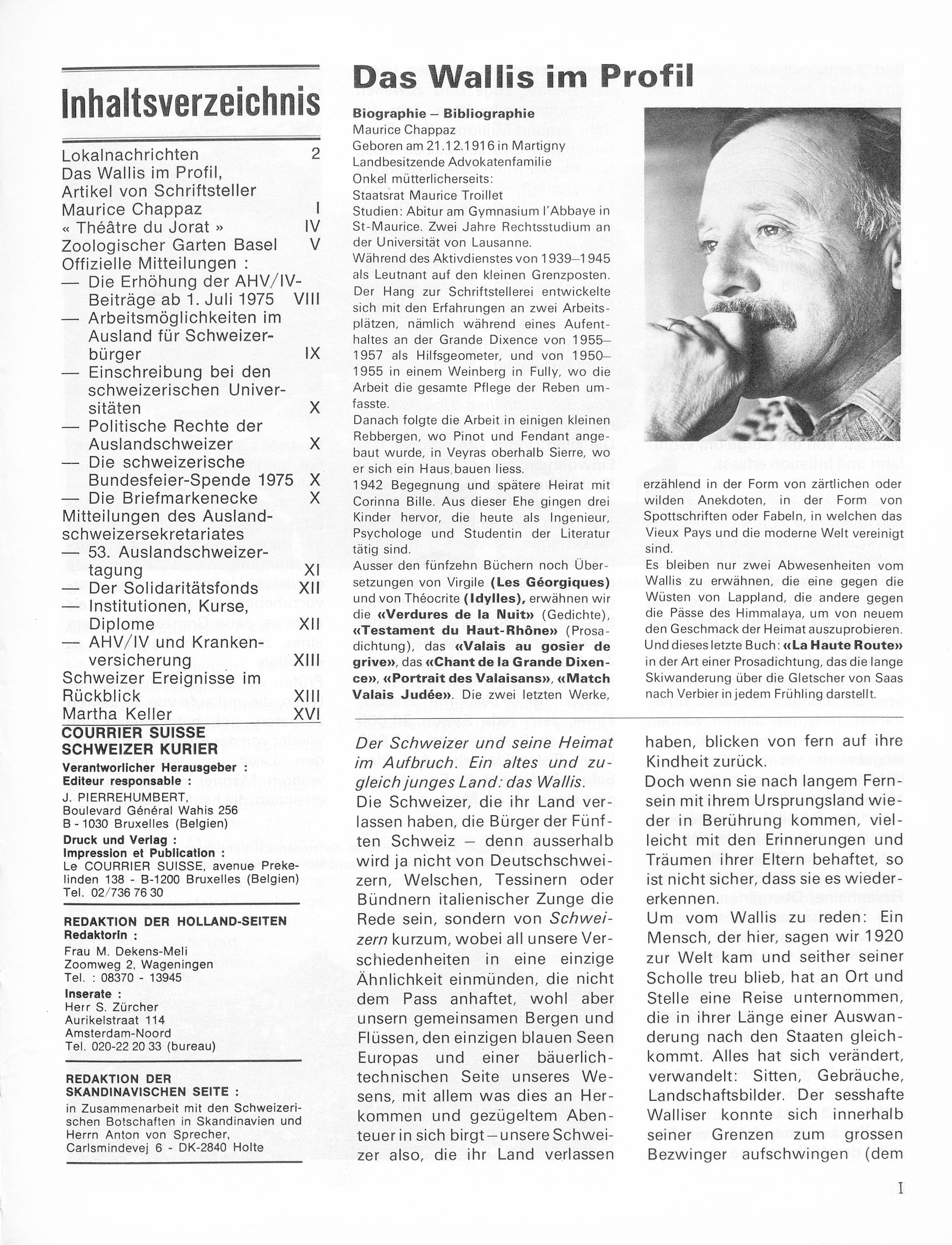}
\caption{Example of document with with a table of contents and an article.}
\label{fig:EP_example3_toc_and_article}
\end{subfigure}

\begin{subfigure}[t]{1.0\linewidth}
\centering
\includegraphics[width=\linewidth]{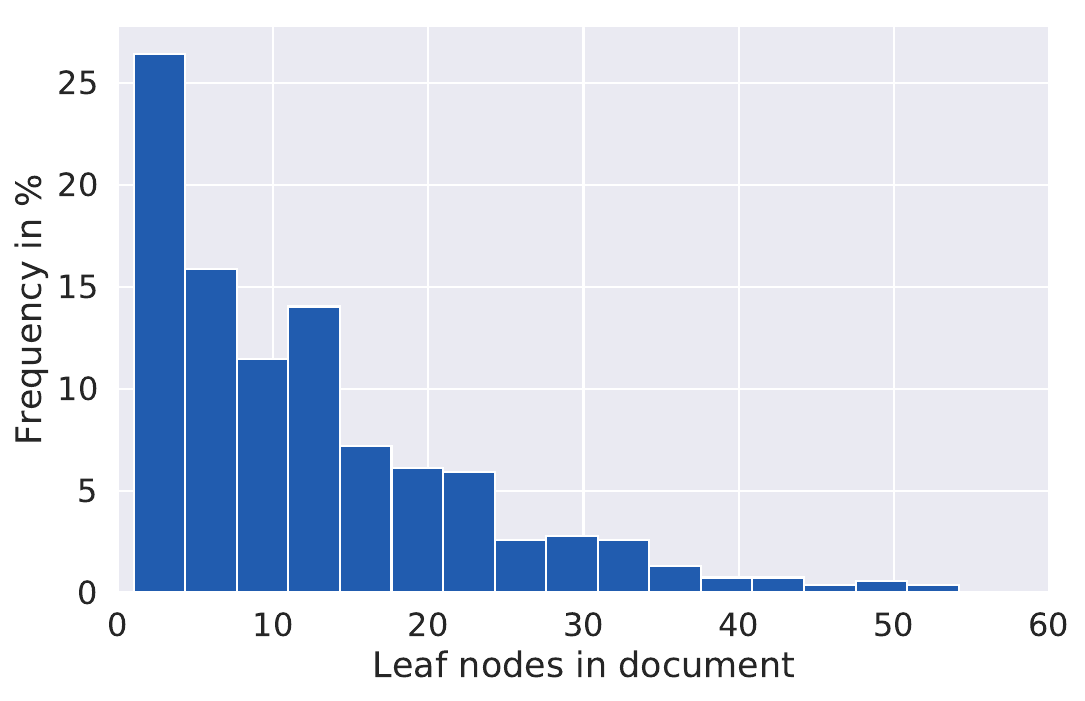}
\caption{Leaf nodes in dataset.}
\label{fig:leaf_histogram}
\end{subfigure}%
\caption{Examples and statistics of documents featured in \datasetEP/.}
\label{fig:EP_example_documents}
\end{figure}

\begin{figure}[tbh!]
\centering
\includegraphics[width=0.5\textwidth]{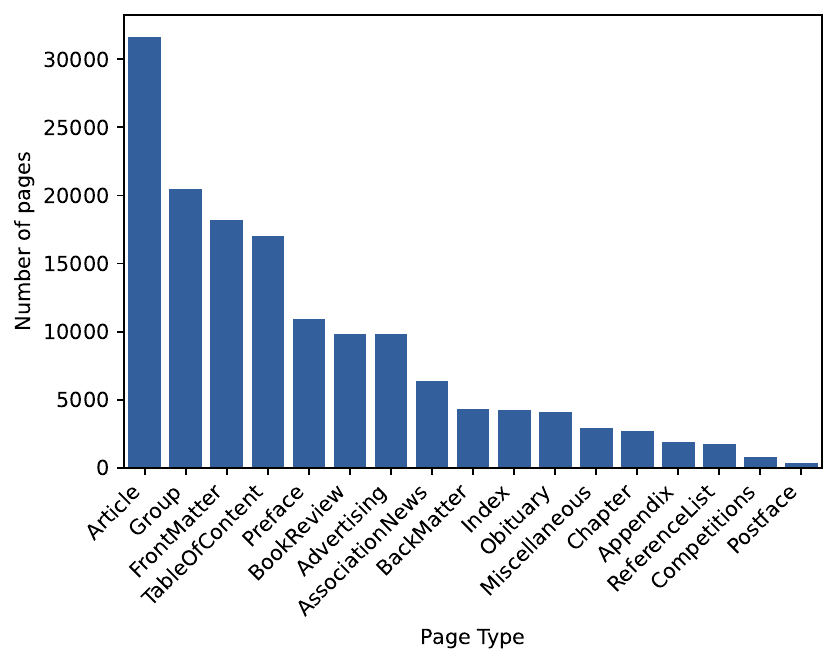}
\caption{Distribution of page types in the \datasetEP/ dataset.}
\label{fig:page_types_bar_chart}
\end{figure}

Figure \ref{fig:page_types_bar_chart} shows the distribution of page types in the \datasetEP/ dataset. %

\textbf{Summary statistics:} Table~\ref{tab:dataset_cat_statistics_manual} lists the semantic categories and their corresponding frequency in the \datasetEP/ dataset. As can be seen from the table, the semantic categories are highly diverse, especially compared to scientific articles \cite{rauschDocParserHierarchicalDocument2021}, implying that \datasetEP/ is a challenging dataset for benchmarking. %

\begin{table}[tbh!]
\footnotesize
\sisetup
  { 
    round-mode         = places      ,
    round-precision    = 2           ,
    table-align-text-post   = false,
  }
\renewrobustcmd{\bfseries}{\fontseries{b}\selectfont}
    \centering
    \begin{tabular}{l S[zero-decimal-to-integer = true, table-format=5.0] S[table-format=2.2, table-omit-exponent,  fixed-exponent = -2] S[table-format=1.2]}
    \toprule
    Category& {Frequency} & {\%} & {Avg. depth}\\
    \midrule
    \textsc{article}              &       651.0          & 0.056875764 & 2.00\\
    \textsc{author}               &       226.0          & 0.019744889 & 3.959537572\\
    \textsc{background fig.}    &        46.0          & 0.004018871 & 5.000000000\\
    \textsc{column}                  &       228.0          & 0.019919623 & 5.977777778\\
    \textsc{text block}        &      1469.0          & 0.128341779 & 4.011547344\\
    \textsc{document root}        &       542.0          & 0.047352787 & 1\\
    \textsc{figure}               &       550.0          & 0.048051721 & 4.069767442\\
    \textsc{figure caption}       &       196.0          & 0.017123886 & 5.057142857\\
    \textsc{figure graphic}       &       516.0          & 0.045081251 & 5.076694915\\
    \textsc{footer}                 &       158.0          & 0.013803949 & 3.000000000\\
    \textsc{footnote}             &        59.0          & 0.005154639 & 4\\
    \textsc{header}                 &       158.0          & 0.013803949 & 3\\
    \textsc{heading}               &      1275.0          & 0.111392626 & 4.011625282\\
    \textsc{item}                 &      1052.0          & 0.091909837 & 5.039177489\\
    \textsc{itemize}              &       144.0          & 0.012580814 & 4.035294118\\
    \textsc{meta}                 &       416.0          & 0.036344575 & 2\\
    \textsc{ordered group}        &      1132.0          & 0.098899179 & 3\\
    \textsc{page nr.}              &       368.0          & 0.03215097  & 3.021917808\\
    \textsc{row}                  &      1460.0          & 0.127555478 & 6.001464344\\
    \textsc{table}                &       124.0          & 0.010833479 & 4.000000000\\
    \textsc{table of content}     &        89.0          & 0.007775642 & 2.000000000\\ 
    \textsc{tabular}              &       124.0          & 0.010833479 & 4.990196078\\
    \textsc{unordered group}      &       463.0          & 0.040450813 & 3\\
    \bottomrule
    \end{tabular}
    \caption{Distribution of semantic categories in \datasetEP/}
    \label{tab:dataset_cat_statistics_manual}
\end{table}

The distribution of leaf nodes per document in Fig.~\ref{fig:leaf_histogram} further emphasizes the complexity of our dataset due to the deeply nested structure.

\section{Results}

\subsection{Qualitative Evaluation}

We performed a qualitative evaluation to demonstrate the effectiveness of \sysdgg/ in practice. For this, we randomly sampled a small subset of documents and then compare our system against both DocParser and commercial OCR systems. 

\textbf{Procedure:} We use two state-of-the-art, commercial OCR systems: \textbf{ABBYY} \cite{abbyyABBYYFineReader2023} and \textbf{Adobe Acrobat} \cite{adobeAdobeAcrobatPro2023}. These OCR systems are able to parse content in different ordering (e.g., top-down and left-right) but without hierarchical information. Further, the output of these OCR systems is limited to a small set of semantic entities and does not focus on preserving the hierarchical document structure (i.e., the nesting). For comparison, we perform a page recognition on the evaluation documents and export the parsed pages to HTML files. We then map the generated results and HTML tags onto the closest matching semantic entities used by \sysdgg/ for comparison. For instance, text regions that are wrapped by a heading tag in HTML are shown as a \textsc{header} bounding box in our qualitative evaluation, while text regions enclosed by a \textsc{aside} entity are assigned the category \textsc{unordered group}. We manually specify the input language (e.g., English, German) before running the tool, if the tool provides such an option.

\begin{figure*}[tbh!]
    \centering
    \begin{subfigure}[b]{0.19\textwidth}
        \centering
        
         \includegraphics[width=\textwidth]{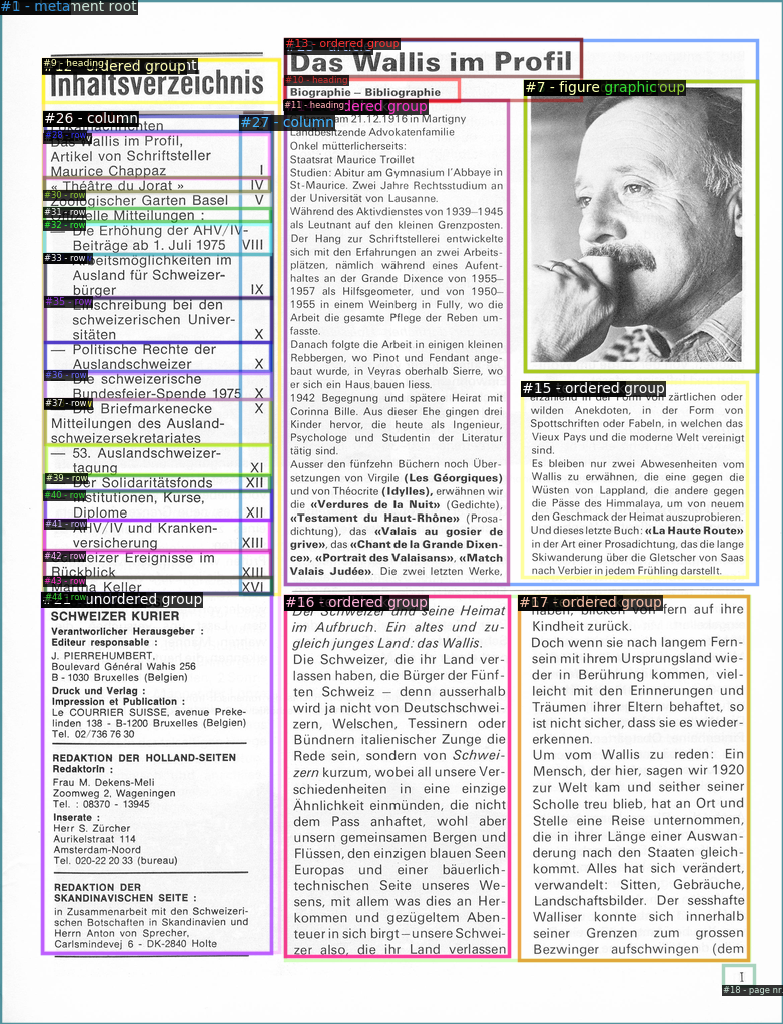}
         
         \caption{Ground-truth.}
         \label{subfig:article_and_toc_gt}
    \end{subfigure}
    \hfill
    \begin{subfigure}[b]{0.19\textwidth}
        \centering
        
         \includegraphics[width=\textwidth,frame]{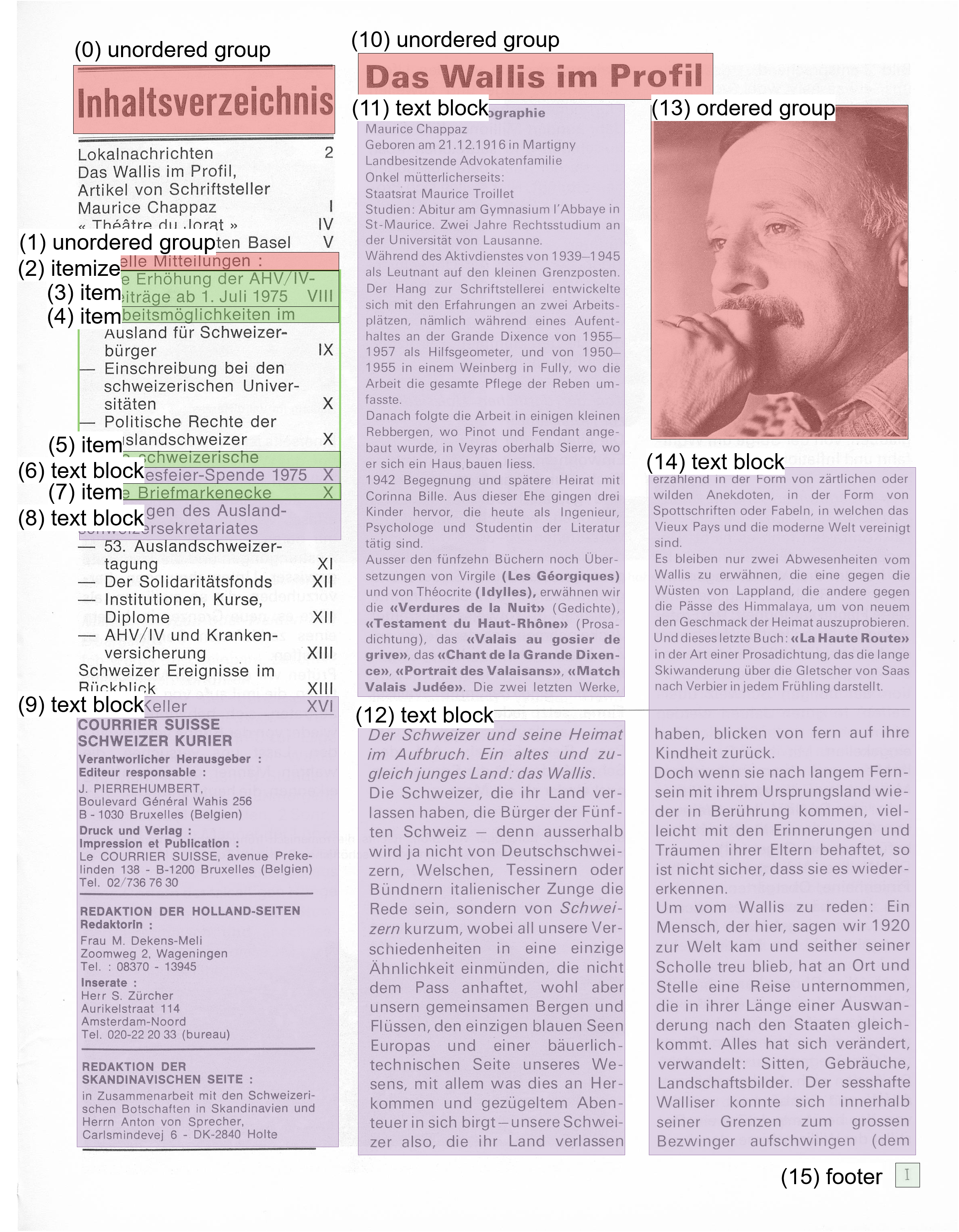}
         
         \caption{ABBYY \cite{abbyyABBYYFineReader2023}.}
         \label{subfig:article_and_toc_abbyy_bboxes}
    \end{subfigure}
    \hfill
    \begin{subfigure}[b]{0.19\textwidth}
        \centering
        
         \includegraphics[width=\textwidth,frame]{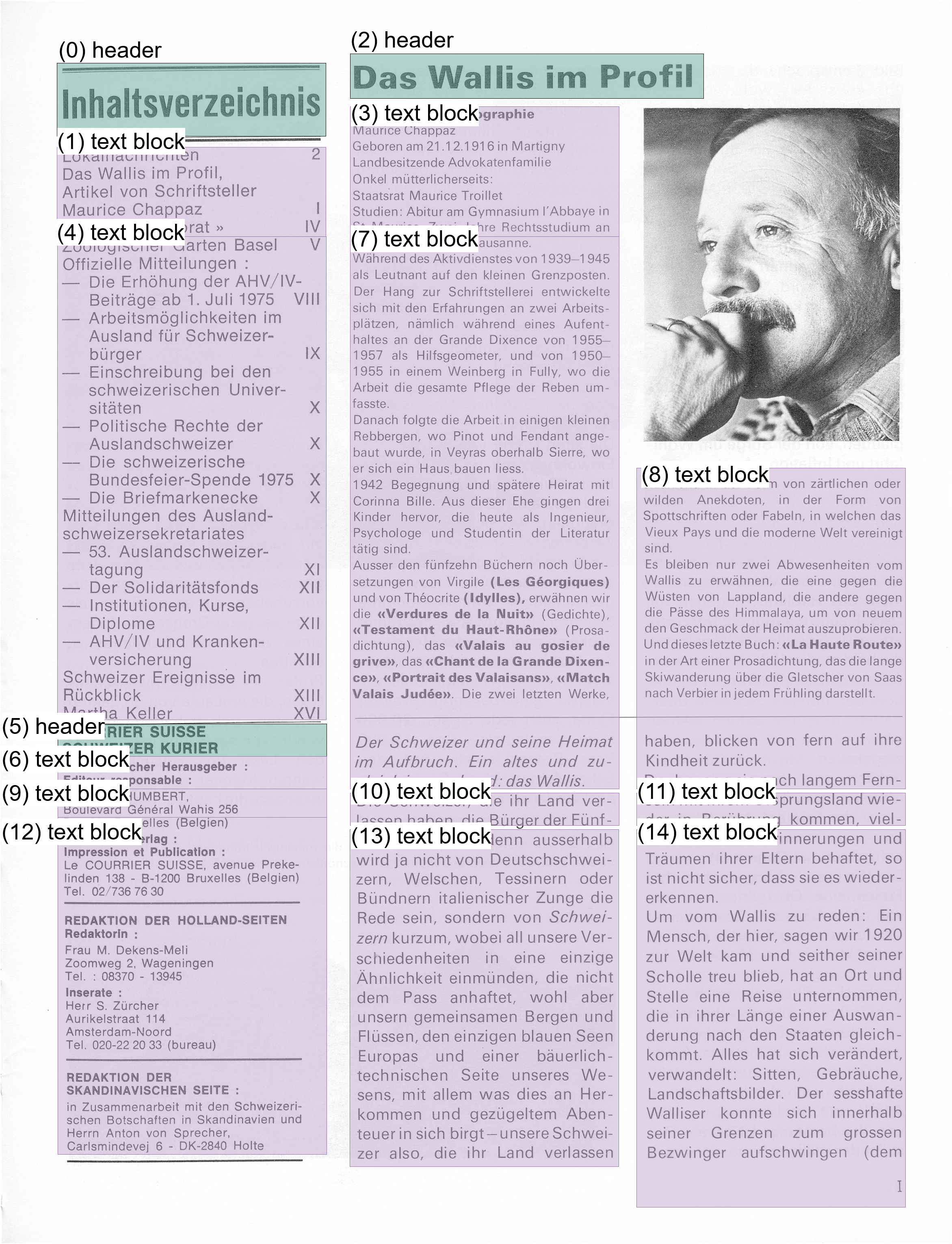}
         
         \caption{Adobe Acrobat \cite{adobeAdobeAcrobatPro2023}.}
         \label{subfig:article_and_toc_acrobat_bboxes}
    \end{subfigure}
    \hfill
    \begin{subfigure}[b]{0.19\textwidth}
        \centering
        
         \includegraphics[width=\textwidth]{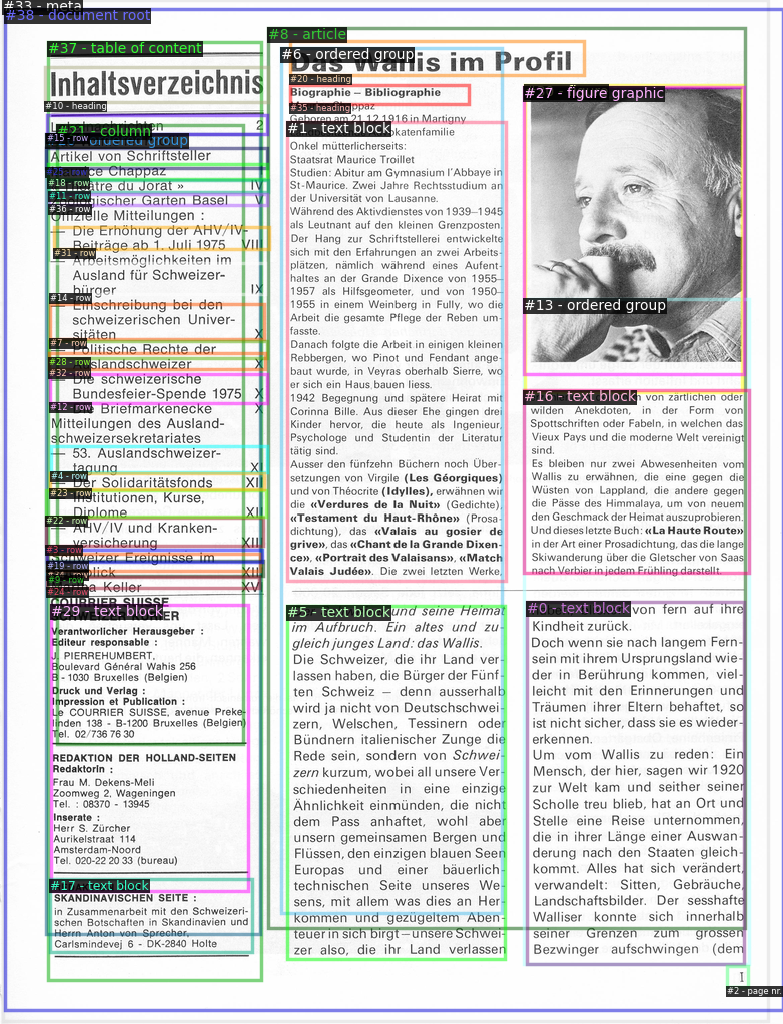}
         
         \caption{DocParser \cite{rauschDocParserHierarchicalDocument2021}.}
         \label{subfig:article_and_toc_legacy_bboxes}
    \end{subfigure}
    \hfill
    \begin{subfigure}[b]{0.19\textwidth}
        \centering
         \includegraphics[width=\textwidth]{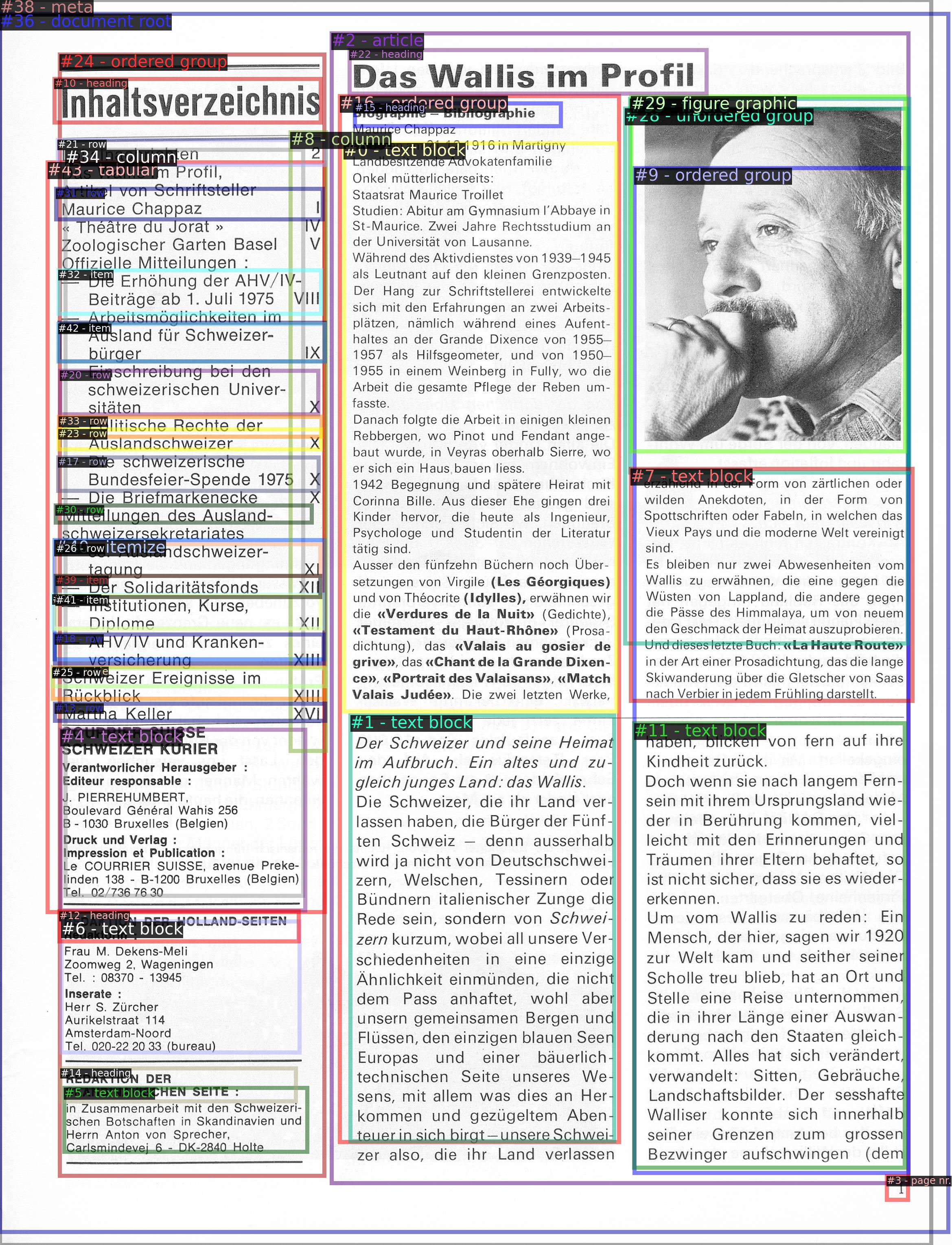}
         \caption{\sysdgg/ (ours).}\label{subfig:article_and_toc_docparserv2_bboxes}
    \end{subfigure}
    \begin{subfigure}[b]{0.19\textwidth}
        \centering
        
         \includegraphics[width=\textwidth]{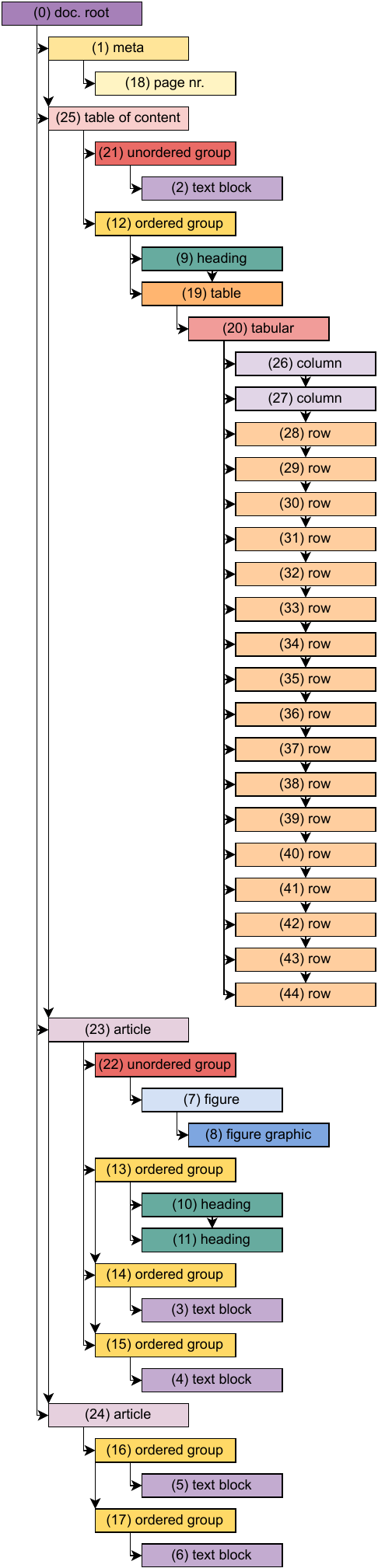}
         \caption{Ground-truth.}
         \label{subfig:article_and_toc_gt_tree}
    \end{subfigure}
    \hfill
    \begin{subfigure}[b]{0.19\textwidth}
        \centering
        
         \includegraphics[width=0.9\textwidth]{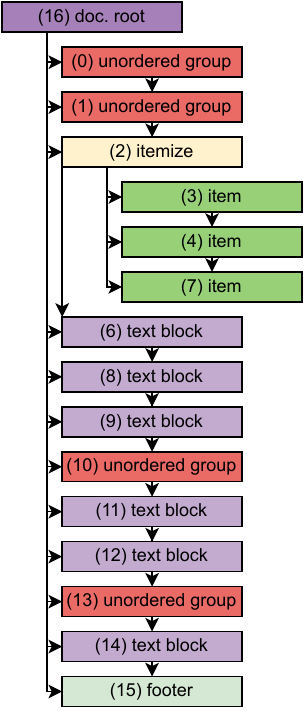}
         
         \caption{ABBYY \cite{abbyyABBYYFineReader2023}.}
         \label{subfig:article_and_toc_abbyy_tree}
    \end{subfigure}
    \hfill
    \begin{subfigure}[b]{0.19\textwidth}
        \centering
        
         \includegraphics[width=.8\textwidth]{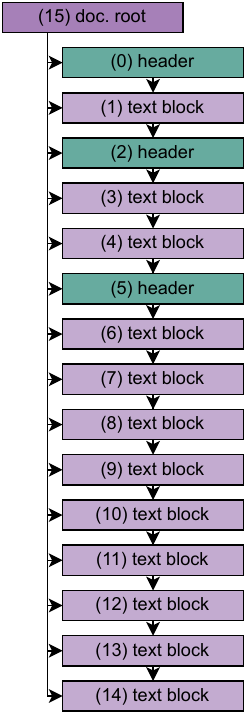}
         
         \caption{Adobe Acrobat \cite{adobeAdobeAcrobatPro2023}.}
         \label{subfig:article_and_toc_acrobat_tree}
    \end{subfigure}
    \hfill
    \begin{subfigure}[b]{0.19\textwidth}
        \centering
         \includegraphics[width=.61\textwidth]{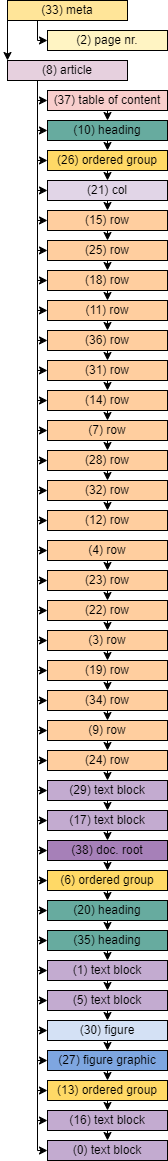}
         \caption{DocParser \cite{rauschDocParserHierarchicalDocument2021}.}
         \label{subfig:article_and_toc_legacy_tree}
    \end{subfigure}
    \hfill
    \begin{subfigure}[b]{0.19\textwidth}
        \centering
         \includegraphics[width=0.92\textwidth]{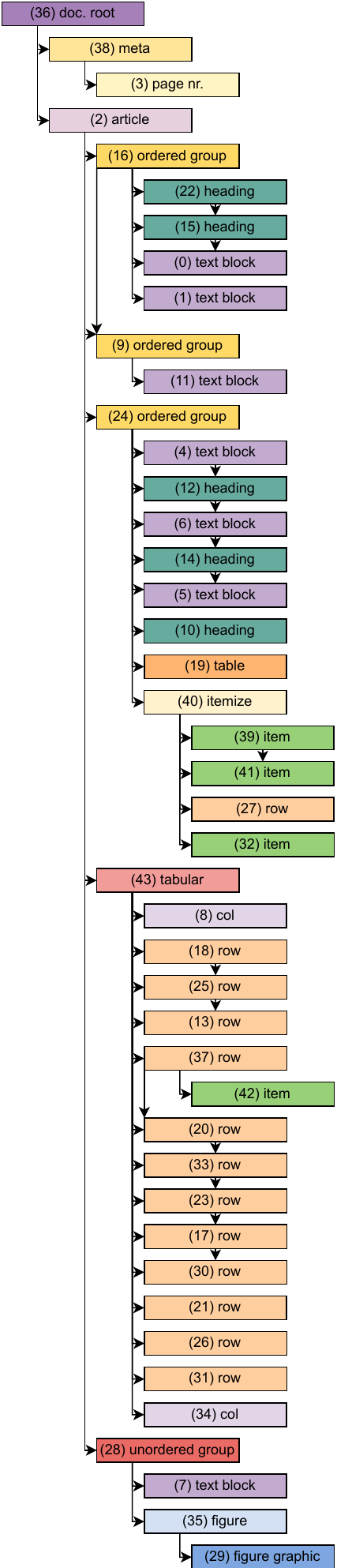}
         \caption{\sysdgg/ (ours).}
         \label{subfig:article_and_toc_docparserv2_tree}
    \end{subfigure}
\caption{Qualitative evaluation comparing the parsed hierarchical document structure by different systems. The document is characterized by complex, hierarchical structure. Top: entity recognition; bottom: hierarchical structure.}
\label{fig:qual_eval_1_toc_and_article}
\end{figure*}

\textbf{Results:} \Cref{fig:qual_eval_1_toc_and_article} shows a representative example.
The unedited input images are shown in \Cref{fig:EP_example3_toc_and_article}
Overall, we find that, even for F1 scores in the order of $\sim$0.5, the final document structure is typically very accurate. While even minor discrepancies or ambiguities between the predicted entities and the ground-truth may lead to notable drops in F1 scores, their overall similarity is still large.

\section{Querying}

\subsection{hOCR Querying}

We further support direct querying of hOCR files for downstream tasks as follows. To this end, we introduce an extended \sysdgg/ syntax for queries using XPath (XML Path Language).

We provide example queries to underline the functionality:
\begin{itemize} \raggedright

\item Using the \sysdgg/ document structure in the enriched hOCR files allows more complex queries such as \lstinline{//div[dsg_cat="orderedgroup"]/*/div[dsg_cat="heading"]/span[@text="results"]/..}\ . This query returns all \textsc{heading} entities that contain the word ``results'' and are a child of an \textsc{ordered group} entity.
\item We enable queries on sequential order, e.g., written as \lstinline{followedby(//div[dsg_cat="heading"], //div[dsg_cat="textblock"])}. This query returns all \textsc{text block} entities that follow a \textsc{heading}. 

The \lstinline{followedby(.,.)} method takes two lists of \sysdgg/ XML nodes or two XPaths as input and returns all nodes from the second list that follow a node from the first list.
\end{itemize}

\subsection{Examples}
We demonstrate an exemplary query on real data (Fig.~\ref{subfig:article_and_toc_docparserv2_tree}) as follows:

\begin{lstlisting}[style=pythonstyle, language=Python] 
>>row_child_of_tabular_and_containing_diplome=root_hocr.xpath('//div[@dsg_class="tabular"]/*/div[@dsg_class="row"]/span[text()="Diplome"]/..')
>>print(entity_child_texts(row_child_of_tabular_and_containing_diplome))
["Institutionen,", "Kurse,", "Diplome,", "XII"]
>>headings=root_hocr.xpath('//div[@dsg_class="heading"]')
>>print_heading_text(headings[:2])
[["Das Wallis im Profil"], ["Biographie", "-", "Bibliographie", "Maurice", "Chappaz"]]
>>textblock_after_biblio=followedby('//div[@dsg_cat="heading"]/span[text()="Biographie"]/..', '//div[@dsg_cat="contentblock"]', root_hocr)
>>print(entity_child_texts(textblock_after_biblio)[:5])
["Geboren", "am", "21.12.191", "6", "in", "Martigny"]
\end{lstlisting}

\noindent

\end{appendices}

\end{document}